\documentclass{sigkddExp}

\usepackage[skip=0pt]{caption}
\usepackage{subcaption}
\usepackage{tabularx}
\usepackage{graphicx}
\usepackage{booktabs} % For formal tables
\usepackage{xcolor}
\usepackage[normalem]{ulem}
\usepackage{paralist}
\usepackage{xspace}
\usepackage{multirow}
\usepackage{siunitx}
\usepackage{bbm}
\usepackage{hyperref}
\usepackage{xfrac}
\usepackage{amsmath}
\usepackage{scalerel}
\usepackage{balance}

\usepackage{color-edits}
\addauthor[Rayid]{rg}{blue}
\addauthor[Kit]{kr}{red}
\addauthor[Hemank]{hl}{cyan}
\addauthor{add}{RedViolet}

\newcommand{\specialcell}[2][c]{%
  \begin{tabular}[#1]{@{}c@{}}#2\end{tabular}}

\begin{document}
%
% --- Author Metadata here ---
% -- Can be completely blank or contain 'commented' information like this...
%\conferenceinfo{WOODSTOCK}{'97 El Paso, Texas USA} % If you happen to know the conference location etc.
%\CopyrightYear{2001} % Allows a non-default  copyright year  to be 'entered' - IF NEED BE.
%\crdata{0-12345-67-8/90/01}  % Allows non-default copyright data to be 'entered' - IF NEED BE.
% --- End of author Metadata ---

%\title{An Empirical Comparison of Bias Reduction Techniques on Real-World Problems}
\title{An Empirical Comparison of Bias Reduction Methods on Real-World Problems in High-Stakes Policy Settings}
\numberofauthors{3}
\author{
%% e-mail address with \email.
\alignauthor Hemank Lamba \\
       \affaddr{Carnegie Mellon University}\\
       \affaddr{Pittsburgh, PA}\\
       \email{hlamba@andrew.cmu.edu}
\alignauthor Kit T. Rodolfa\textsuperscript{\scaleto{*}{4pt}}\\
% \alignauthor Kit T. Rodolfa\authornote{These authors contributed equally.}\\
       \affaddr{Carnegie Mellon University}\\
       \affaddr{Pittsburgh, PA}\\
       \email{krodolfa@cmu.edu}
\alignauthor Rayid Ghani\titlenote{These authors contributed equally.}\\
% \alignauthor Rayid Ghani\authornotemark[1]\\
       \affaddr{Carnegie Mellon University}\\
       \affaddr{Pittsburgh, PA}\\
       \email{rayid@cmu.edu}
}

%\date{March 15, 2021}
\maketitle
\begin{abstract}
Applications of machine learning (ML) to high-stakes policy settings --- such as education, criminal justice, healthcare, and social service delivery --- have grown rapidly in recent years, sparking important conversations about how to ensure fair outcomes from these systems. The machine learning research community has responded to this challenge with a wide array of proposed fairness-enhancing strategies for ML models, 
% which can be broadly categorized by the point in the ML pipeline to which they apply: 1) pre-processing methods that adjust the training data, 2) in-processing methods that add equity constraints in model training, and 3) post-processing methods that act on trained models or scores. 
but despite the large number of methods that have been developed, little empirical work exists evaluating these methods in real-world settings. Here, we seek to fill this research gap by investigating the performance of several methods that operate at different points in the ML pipeline across four real-world public policy and social good problems. 
%settings in which ``equality of opportunity'' (as measured by True Positive Rate disparity) is an appropriate conceptualization of fairness. 
Across these problems, we find a wide degree of variability and inconsistency in the ability of many of these methods to improve model fairness, but post-processing by choosing group-specific score thresholds consistently removes disparities, with important implications for both the ML research community and practitioners deploying machine learning to inform consequential policy decisions.
\end{abstract}

\section{Introduction}
There has been a recent increase in the use of machine learning models to support decisions in high stakes domains with societal impact, including informing bail decisions~\cite{chouldechova2017fair,skeem2016risk,angwin2016machine}, hiring~\cite{raghavan2020mitigating}, healthcare delivery~\cite{obermeyer2019dissecting,ramachandran2020predictive} and social service interventions~\cite{bauman2018reducing,chouldechova2018case,potash2015predictive}. These decisions affect critical aspects of people's lives and if not done responsibly, can hurt already vulnerable and historically-disadvantaged communities. This combination of increased use, increased potential for improving social outcomes, and increased risk of harm has prompted questions from  researchers, policymakers, citizens, and the media about the role these models can play in exacerbating (or reducing)  existing inequities \cite{howard2018ugly,osoba2017intelligence,mehrabi2019survey,caton2020fairness}, giving rise to a growing area, FairML, focused on dealing with issues of bias and fairness in building and using machine learning systems. 
FairML research has grown to span issues around defining bias in machine learning models, enumerating a variety of metrics that can be used to measure model bias \cite{Verma2018FairnessExplained}, detecting instances of it through audit tools \cite{saleiro2018aequitas,bellamy2019ai}, and methods for reducing (or mitigating the impact of) bias in ML models \cite{caton2020fairness}. In this work, we focus on bias reduction methods, which can be broadly categorized into three groups based on the stage of analysis at which they are applied:
\begin{enumerate}
\item \textit{Pre-processing methods} typically involve changing the data in some manner \textit{before} building models.
\item \textit{In-processing methods} typically involve using ML models/methods that are explicitly designed to deal with bias \textit{in the process of building models}, such as using regularization approaches.
\item \textit{Post-processing methods} typically involve adjusting scores, thresholds, or model selection \textit{after the model predictions} have been generated.
\end{enumerate}

Despite active research and the development of several new methods in the area of FairML in recent years, there has been a lack of extensive empirical evaluation across them to assess their effectiveness on real-world problems and data. The majority of this research has typically focused on achieving abstract, general-purpose definitions of fairness, and evaluated on benchmark data sets (such as the adult data set \cite{Dua:2019}) or limited data sets (such as  COMPAS \cite{larson2016we}). While that is a reasonable starting point, benchmark data sets often do not reflect the richness, nuances, and constraints of real-world problems, making it unclear for both researchers and practitioners how to assess the applicability and effectiveness of these methods or make decisions around which ones to use under given circumstances in real-world situations. 

In this paper, we attempt to fill this empirical gap by presenting a \textbf{comprehensive empirical evaluation} of different bias reduction strategies over \textbf{four real-world problems} that come from public policy and social good settings. We want to emphasize that real-world problems are not just data sets but rather a combination of business/policy problems, the corresponding machine learning formulation and evaluation metrics, an extensive set of features generated in the feature engineering process, a large and varied set of ML models and hyperparameters, and a validation methodology and metric(s) that mirrors the deployment scenario. To that end, we describe the analytical formulation for each of these problems, the feature engineering process, parameters of temporal model selection using a wide variety of ML models and hyperparameters, and then evaluate the effectiveness of a variety of bias reduction strategies in reducing specific disparities while preserving as much of the original evaluation metric of interest as possible. We believe that this paper not only fills a critical gap today for researchers and practitioners of FairML but also provides a framework for researchers proposing new methods to follow when reporting the effectiveness of their work.

\section{Related Work}
The focus of this paper is not on the entire process of building ML systems that lead to fair and equitable outcomes but more narrowly on methods that are used to reduce the bias in the predictions of ML models. With that focus, as mentioned earlier, bias reduction methods can be categorized broadly into three categories, based on the phase of the analysis pipeline to which they are applied: (a) Pre-processing, (b) In-processing and, (c) Post-processing. 

\subsection{Pre-processing}

Pre-processing approaches assume that the bias in the ML models is caused by certain variables in the data or by the distribution of the data being used to train and validate the ML models. Most of the pre-processing approaches thus try to modify the data by either removing the sensitive variable (gender or race for example) or by changing the data distribution (with respect to the sensitive variable) by sampling.

\textbf{Omission of sensitive variables} has been widely explored in the past~\cite{hajian2012methodology, salimi2019interventional}. This approach is based on the assumption that if machine learning model is not given the protected variable as a feature, the model that is trained will not be dependent on the protected variable,  making the model unbiased. Unfortunately, this assumption is often overly-optimistic (and violated) in real-world problems where several other features, including ones relevant to the prediction problem, may be strongly correlated with the protected attribute.
Recent work has described how omission of sensitive variables for training models often may not affect bias reduction (or even increase biases) despite decreasing model accuracy~\cite{calmon2017optimized,kamiran2012data,dwork2012fairness}. Despite these well-documented limitations, we included this strategy in the present exploration of fairness-enhancing methods because this notion of ``fairness through unawareness'' nevertheless persists and has commonly been posited by policymakers, decision-makers (in governments, non-profits, and corporations), and students we have worked with. Notably, other researchers have proposed more nuanced approaches to modifying the input data to remove correlations with the protected attribute in addition to the attribute itself. Although we do not explore this direction for pre-processing here, we refer the reader to \cite{feldman2015certifying} for an example of this approach in the context of disparate impact as a measure of fairness.

\textbf{Resampling} involves modifying the distribution of the training data by either over- or under-sampling examples to reduce disparities when the modified data is used in model training. Calders et al.~\cite{calders2010three} explored three different sampling techniques to fix existing bias in the data distribution to ensure that a model (in their case, Naive Bayes) trained on the modified data is more fair. Similarly, Iosifidis et al.~\cite{iosifidis2019fae} used clustering across sensitive attribute and labels to come up with representative training data to train models, and Kamiran et al. \cite{kamiran2012data} explored multiple techniques involving sampling and re-weighing of training instances as pre-processing steps before applying machine learning models. Other popular preprocessing techniques involve relabelling and perturbation \cite{balancinglabels}, details of which we omit from the paper.

\subsection{In-processing}
In-processing bias reduction methods generally include regularization or constrained optimization approaches to account for fairness metrics while solving their underlying classifier's optimization problem. Regularization adds penalty terms to the objective function of the classifier such that it is penalized for unfair solutions, whereas constrained optimization generally introduces fairness as a hard constraint in order to directly reject solutions that fail to satisfy fairness criteria. Kamishima et al.~\cite{kamishima2011fairness} proposed a regularization technique that uses mutual information of the sensitive attribute and prediction class, penalizing any increase in conditional probability on a specific subgroup. Zafar et al. \cite{Zafar2017FairnessClassification} extended on this work by introducing fairness constraints into the objective function of the underlying classifier. One challenge faced by these approaches, however, is that these constraints often yield a non-convex objective function, making the optimization problem inherently difficult. To address this issue, Zafar proposed an efficient method for solving the resulting non-convex formulation. Similar techniques for different fairness metrics and even general classes of metrics have also been proposed in the literature~\cite{celis2019classification}. Jiang et al. proposed an approach that minimizes Wasserstein-1 distances between classifier output and sensitive information~\cite{jiang2020wasserstein}. Heidari et al. proposed a Rawlsian concept of fariness that can be introduced as a constraint into any convex loss-minimization algorithm~\cite{heidari2018fairness}. Similar methods have also been extended to neural-network based models~\cite{manisha2018neural} as well as decision trees~\cite{aghaei2019learning}. 
% There are other lot of work that we have omitted from this section but we refer readers to for a much more comprehensive review.

\subsection{Post-processing}
Post-processing methods are generally agnostic to the machine learning models used, and modify the outputs to improve fairness in predictions or classifications. This involves training meta-models with fairness constraints~\cite{celis2019classification,dwork2018decoupled} or directly thresholding or modifying model scores to improve fairness ~\cite{hardt2016equality, rodolfa2020case}. Hardt et al. proposed methods for \textbf{direct post-hoc adjustments to scores} (or binary predicted classes) from trained classifiers to achieve either equalized odds or equality of opportunity by choosing group-specific thresholds that meet these fairness goals~\cite{hardt2016equality}. Recently, we have extended on this work, applying similar methods across a number of policy contexts and finding little or no trade-off in model accuracy in doing so~\cite{rodolfa2020case, rodolfa2020machine}.

Another fairness-enhancing strategy that can be applied on top of a range of underlying machine learning methods involves \textbf{decoupling the training or selection of classifiers}, as proposed by Dwork and colleagues~\cite{dwork2018decoupled}. This approach starts from the hypothesis that a model trained to do well on the entire population might not fully capture differences in predictiveness of features or other important patterns across groups and posits that training separate models for each protected group might better pick up on these nuances. Because fully decoupling the models might significantly reduce the available training data (particularly for small groups), they also suggest exploring different levels of transfer learning between groups, giving a relative weight to training examples from the protected group or rest of the population (so, at the other extreme, one might train models across the full population, but select best-performing models for each group rather than a single overall model).

Other authors, including Celis et al~\cite{celis2019classification} as well as Menon and Williamson~\cite{menon2018cost}, have proposed methods that perform a \textbf{constrained optimization to train a meta-model} to improve the fairness of a prediction score generated by a model. These methods seem particularly useful where membership in the protected groups is not known apriori but can be estimated (for instance,~\cite{celis2019classification} describes estimating a joint probability distribution over outcomes and sensitive attributes). However, when group membership is known, these methods will generally result in stretching or shifting within-group score distributions without reordering in a manner equivalent to choosing separate thresholds for each group (for more detail, see our discussion in the supplemental materials from~\cite{rodolfa2020machine}).

Finally, and perhaps most simply, fairness can be incorporated into the process of \textbf{model selection}. After training a large set of different model types and hyperparameter values, the validation set performance of these different trained models can be assessed both in terms of traditional accuracy metrics (such as AUC-ROC, precision@k, or other confusion matrix based metrics) as well as fairness metrics appropriate to the context. Choosing a model to deploy then becomes an optimization problem over two dimensions, with a Pareto frontier reflecting a menu of potential trade-offs between these two goals of accuracy and fairness~\cite{williams2018model}. In practice, the trade-offs presented by this frontier might be a function of inherent properties of the data and problem as well as the extent to which the grid search that was performed covers the possible space of model types and hyperparameters. Although relatively straightforward in nature and implementation, relying entirely on model selection is somewhat arbitrary as it relies entirely on finding a model specification that performs well on both fairness and accuracy metrics without taking active steps to ensure or improve fairness.

\section{Comparison Setup}
This section describes our setup to conduct the empirical evaluation across bias reduction methods. We describe the specific methods we chose to compare, the policy contexts for the problems we use to conduct that empirical evaluation, and the specific experimental setup for each real-world problem (the data used, features generated, models built, evaluation metric, protected group, and bias metric).

\subsection{Methods to Compare}
While a large number of bias reduction methods exist in each category we describe in Section 2 (Pre-processing, In-processing, and Post-processing), in this paper, we focus on a few representative methods from each category to compare with each other. The methods chosen for this study are described below.

\subsubsection{Pre-Processing Methods}

\textbf{Removing the Protected Attribute}: For each problem domain, we define a set of protected attributes and remove those from the data before performing any ML modeling.

\textbf{Sampling}:  We apply sampling to our training sets with respect to the protected group in three ways: a) changing the marginal distribution of the protected and non-protected subgroups, b) changing the label distribution within the protected and non-protected subgroups, and c) changing both simultaneously. Here, we implemented the six sampling strategies described in Table \ref{table:sampling} reflecting a set of reasonable a priori hypothesis about how these distributions in the training data might influence model fairness.

To formalize our sampling approaches, we define $Protected$ as the protected value/group (such as Race=Black) and $NonProtected$ as the set of values that are considered Non-Protected (such as Race=White). The (binary) label variable is represented as $Y$ with values 0 and 1. $P^0(\cdot)$ represents a probability distribution in the original dataset and $P'(\cdot)$ represents a probability distribution after resampling. With those definitions, each of our three sampling settings are:

\textbf{(A)} Balances the data by changing the ratio of Protected to Non-Protected while preserving the original label distribution within each group.

The goal is to achieve:
\[
\frac{P'(NonProtected)}{P'(Protected)}  = \alpha
\]
while preserving the original label distribution within $Protected$ and $NonProtected$ such that
\begin{align}
 P'(Y=1 \mid NonProtected) & = P^0(Y=1 \mid NonProtected)    \notag \\
\text{and  } P'(Y=1 \mid Protected) & = P^0(Y=1 \mid Protected) \notag
\end{align}

In Table \ref{table:sampling}, Strategy 1 uses this approach with $\alpha = 1$.

\textbf{(B)} Balances the label distribution across each subgroup: Protected and NonProtected. The goal is to achieve:
\begin{align}
P'(Y=1 \mid NonProtected) &= \beta_{NP} \notag \\
P'(Y=1 \mid Protected) &= \beta_P \notag \\
\text{such that~} \frac{\beta_{NP}}{\beta_P} &= \gamma \notag
\end{align}

while preserving the original marginal distributions for Protected and NonProtected such that:
\begin{align}
    P'(NonProtected) & = P^0(NonProtected)  \notag \\ 
    \text{and  } P'(Protected) & = P^0(Protected) \notag
\end{align}

In Table \ref{table:sampling}, Strategy 2 uses this approach (with $\beta_P = \beta_{NP} = 0.5$ and $\gamma = 1$), as does Strategy 3 (with $\beta_P = \beta_{NP} = P^0(Y=1 \mid NonProtected)$ and $\gamma = 1$) and Strategy 4 (with $\beta_{NP} = P^0(Y=1 \mid NonProtected)$ and $\beta_P = 0.5$).

\textbf{(C)} Adjusts the marginal distribution of Protected and NonProtected as well as the label distributions by setting $\alpha$, $\beta_P$, $\beta_{NP}$, and $\gamma$ as described above.

In Table \ref{table:sampling}, Strategy 5 uses this approach (with $\alpha = 1$ and $\beta_P$ = $\beta_{NP} = 0.5$) as does Strategy 6 (with $\alpha=1$, and $\beta_P = \beta_{NP} = P^0(Y=1 \mid NonProtected)$).

%\krcomment{Maybe too long winded spelled out like this?}
%\hlcomment{I think this looks good}

Note that in each strategy, in order to balance two distributions, we can either \textit{undersample} from the majority distribution or \textit{oversample} from the minority distribution. In case of oversampling, we randomly sample (with duplicates allowed) to generate more examples,\footnote{For oversampling, we do not make use of methods such as SMOTE~\cite{chawla2002smote} as each feature might have a specific set of constraints and this method does not take into account the overall joint distribution.} increasing the total number of examples as little as possible while achieving the desired distributions. When undersampling, we remove as few examples as possible in order to achieve the desired distributions. Also note that we only sample in each training set while keeping the distribution of the validation sets the same as in the original data.

\begin{table}[!hbtp]
    \centering
    \caption{Sampling strategies used in this study.}
    \begin{tabular}{cccc}
    \toprule
         \textbf{} & \textbf{\specialcell{Ratio: Protected\\to Non-Protected}} & \textbf{\specialcell{Label Dist.\\Protected}} & \textbf{\specialcell{Label Dist.\\Non-Protected}} \\
         \midrule
         1 & 1:1 & Original & Original \\ \midrule
         2 & Original & 50-50 & 50-50  \\  \midrule
         3 & Original & \specialcell{Same as\\ Non-Protected} & Original   \\  \midrule
         4 & Original & 50-50 & Original \\ \midrule
         5 & 1:1 & 50-50 & 50-50  \\  \midrule
         6 & 1:1 & \specialcell{Same as\\ Non-Protected} & Original \\
    \bottomrule
    \end{tabular}
    \label{table:sampling}
\end{table}

% Kit: Removing the "Strategy" column as I don't think we use it anywhere in the text and the names don't feel too intuitive. Here's the original if we want it back:

% \begin{table*}[!hbtp]
%     \centering
%     \caption{Different sampling strategies.}
%     \begin{tabular}{ccccc}
%     \toprule
%          \textbf{Strategy Id} & \textbf{Strategy} & \textbf{\specialcell{Ratio of Protected\\to Non-Protected}} & \textbf{\specialcell{Label Dist.\\Protected}} & \textbf{\specialcell{Label Dist.\\Non-Protected}} \\
%          \midrule
%          1 & 1-Orig-Orig & 1:1 & Original & Original \\ \midrule
%          2 & Orig-50-50 & Original & 50-50 & 50-50  \\  \midrule
%          3 & Orig-Snop-Orig & Original & Same as Non-Protected & Original   \\  \midrule
%          4 & Orig-50-Orig & Original & 50-50 & Original \\ \midrule
%          5 & 1-50-50 & 1:1 & 50-50 & 50-50  \\  \midrule
%          6 & 1-Snop-Orig & 1:1 & Same as Non-Protected & Original \\
%     \bottomrule
%     \end{tabular}
%     \label{table:sampling}
% \end{table*}

\subsubsection{In-Processing Methods}
In this paper, we focus on in-processing through constrained optimization to reduce model disparities. This approach includes fairness metrics in the objective function and seeks to produce predictions that maximize accuracy while taking fairness into account.

Zafar and colleagues~\cite{Zafar2017FairnessClassification,Zafar2017FairnessMistreatment} proposed a constrained optimization method centered on a fairness notion they described as ``disparate mistreatment.'' A model can be said to have disparate mistreatment when misclassification rate for the protected and non-protected group are different, and their work described optimization problems using either False Positive Rate (FPR) or False Negative Rate (FNR) as a measurement of misclassification. Formally, this optimization problem (for FNR) is defined by:
\begin{equation*}
\begin{aligned}
\min {L}(\theta)    \\
\textrm{s.t. } P(\hat{y} \neq y \mid z=0, y=1) &- P(\hat{y} \neq y \mid z = 1, y=1) \leq \epsilon    \\
P(\hat{y} \neq y \mid z=0, y=1) &- P(\hat{y} \neq y \mid z = 1, y=1) \geq -\epsilon
\end{aligned}
\end{equation*}

where, $L$ is the loss function (over model parameters $\theta$), $\hat{y}$ prediction, $y$ original label, $z$ is the protected attribute, and $\epsilon$ denotes the tolerance boundaries for a fair output.

For our problem settings, we focus on True Positive Rate (TPR) disparities (also referred to ``equality of opportunity'' by Hardt \cite{hardt2016equality}) as the appropriate metric of fairness (see the discussion on problem settings below, as well as in \cite{Rodolfa2020CaseInterventions,rodolfa2020machine}). However, because $TPR = 1-FNR$, we make use of Zafar's method to equalize FNR. In doing so, we used a very small value of $\epsilon = 0.0001$ to find solutions which remove disparities entirely.

Recently, open source toolkits such as FairLearn~\cite{bird2020fairlearn} have also been introduced which try to reduce biases, according to a given metric in classification problems. However, we do not include FairLearn in this study setting because it only generates binary predicted class labels rather than a continuous score. This makes it poorly suited to our problem settings where we focus on choosing the k highest-risk entities for intervention based on an organization's resource constraints (as discussed in more detail below). In other work, we have explored heuristics such as sampling to select top k predictions from the output of FairLearn but found that it performed poorly since it wasn't designed for that purpose ~\cite{saleiro2020dealing}.

\subsubsection{Post-Processing Methods}
We define the post-processing class of methods as any method that is applied once the model has been built, typically in adjusting the scores that the models produced or using different thresholds to create classification decisions. We describe several such methods above and discuss here the methods we explored in the present work.

\textbf{Post-Hoc Adjustments}: Here we expand on some of our recent work~\cite{rodolfa2020case,rodolfa2020machine} using a method to equalize TPRs across groups while keeping the total number of individuals selected constant, reflecting the ``top k'' setting of the policy problems we consider (see the discussion on problem settings below for more details). In short, because TPR increases monotonically with depth in a predicted score, we can find a single solution (up to randomized tie breaking) with equalized TPR across groups by adjusting the score thresholds for each group while keeping the total number of individuals selected constant. In practice, these threshold adjustments are made on the model scores in one validation split (say, at time $t=0$) to decompose the overall number of individuals to select by group,\footnote{For instance, if a program can intervene on 100 individuals, this process might break that down into 75 Black individuals and 25 white individuals. Because score distributions are likely to change over time, group-specific ``top k'' values are used rather than score thresholds to ensure the total number of targeted individuals remains fixed.} then these group-specific target numbers are applied to a subsequent validation set to evaluate how well this fairness-enhancing strategy generalizes into the future. Note that, as mentioned above, some of the meta-model approaches such as those described in~\cite{celis2019classification,menon2018cost} can be shown to be mathematically equivalent to choosing different score thresholds when protected group membership is known (rather than modeled) and a unique equitable solution exists, as is the case here. As such, we don't explore those methods separately from these post-hoc adjustments through group-specific thresholding.

\textbf{Composite Models}: Following the proposal of Dwork and colleagues~\cite{dwork2018decoupled}, we investigated two options for building composite models from models trained or selected for their performance on subgroups. On the one extreme, we simply used the grid of models trained on the full population but performed model selection separately for each subgroup (reflecting the complete transfer learning approach described by Dwork). On the other extreme, we trained separate models just with examples from each subgroup (the fully decoupled approach in Dwork) and added these to the model grid for subgroup-specific model selection. One challenge with implementing these composite models, however, is that the scores from the separate models chosen for different subgroups have not been calibrated and cannot be assumed to be comparable. As such, one needs to determine how to appropriately choose a total ``top k'' set of individuals across these different models. Because we were making use of these composite models in the interest of improving fairness, a natural means of choosing these thresholds was to apply the same method choosing TPR-equalizing thresholds described above. It is somewhat challenging to determine whether fairness improvements seen from these composite strategies are more a result of the group-specific thresholds or decoupling the model building or selection itself. However, one hope here would be that the decoupling should improve the accuracy of model predictions on the subgroups, so success for these methods ideally would show not just similar disparity mitigation to post-hoc adjustments but also improved overall accuracy metrics at the same level of fairness.

\textbf{Model Selection}: As noted above, an additional simple approach that falls under our umbrella of post-processing strategies is to account for fairness metrics in the process of model selection. However, this approach is not only very sensitive to the machine learning method/hyperparameter grid explored but also relies on some degree of luck that specifications with favorable trade-offs will be found. Here, we explored two options by which fairness could be included in the model selection process:
\begin{itemize}
    \item Setting a ``Disparity Constraint'' reflecting a largest acceptable disparity. Here, we only consider models with disparity no higher than a certain value, then choose the model with the highest precision among these. Note that it may be possible that no models have a low enough disparity to meet the criteria, in which case we choose the model closest to this cut-off (making it a soft constraint and guaranteeing a model will always be chosen).
    \item Setting an ``Accuracy Constraint'' reflecting a largest acceptable loss in accuracy to improve fairness. Here, we only consider models with precision@k within a given number of percentage points below the best model, then choose the model with lowest disparity among these. Note that because this constraint is relative to the performance of the most-accurate model, there will always be at least one meeting the criteria, so this is a hard constraint.
\end{itemize}

For each type, we explored eight levels of the constraint, from placing little or no weight on fairness to strongly selecting for fair models. For Disparity Constraints, these included allowing disparities up to 5.0, 2.0, 1.5, 1.3, 1.2, 1.1, 1.05, or 1.0 (that is, exact equity). For the Accuracy Constraints, these included allowing a decrease in precision of up to 0.0, 0.05, 0.10, 0.15, 0.20, 0.25, 0.50, and 0.60 percentage points.

\subsection{Problems, Data, and Experimental Setup}
Our empirical evaluation of these methods was done on three real world problems that we have worked on in collaboration with various government agencies. These span mental health and criminal justice (with Johnson County, Kansas), housing safety inspections (with San Jose, CA), and education outcomes (with the Education Ministry of El Savador). Since the data for these problems is confidential and not available publicly, we also replicate this empirical evaluation on a crowdfunding problem from DonorsChoose\footnote{http://wwww.donorschoose.org} where the data is publicly available. This will allow other researchers and practitioners to replicate our work before applying it to their own problems. In general, these problem settings involve six elements:

\begin{enumerate}
\item{\textbf{Features}: Each project we use in this study went through an extensive feature engineering process. As is typically done in real-world ML systems, the features generated included raw and transformed information about the entities of interest (such as demographics) as well as temporal and spatial aggregations (while respecting temporal boundaries in train and validation sets to avoid leakage).}

\item{\textbf{Label}: In each of the problem domains, the decision on the definition of the label is part of the formulation process and is done in collaboration with the partnering organization. In all of these problems, the label was determined by the occurrence of an event at some point in the future from the time of prediction, for example, an individual being booked into jail in the next 12 months or a crowdfunding project failing to get fully funded in the next 4 months.}

\item{\textbf{Train and Validation Splits}: Since most real-world prediction problems are temporal in nature and violate stationary distribution assumptions, we use temporal validation to split our datasets into train and validation sets~\cite{hyndman2018forecasting}. These train and validation sets are usually temporally sequential in nature, where each candidate model is trained on data from ``past'' data and validated on ``future'' data (see Figure \ref{fig:temporal_validation} for a diagram).}

\begin{figure}
    \centering
    \includegraphics[width=1.0\linewidth]{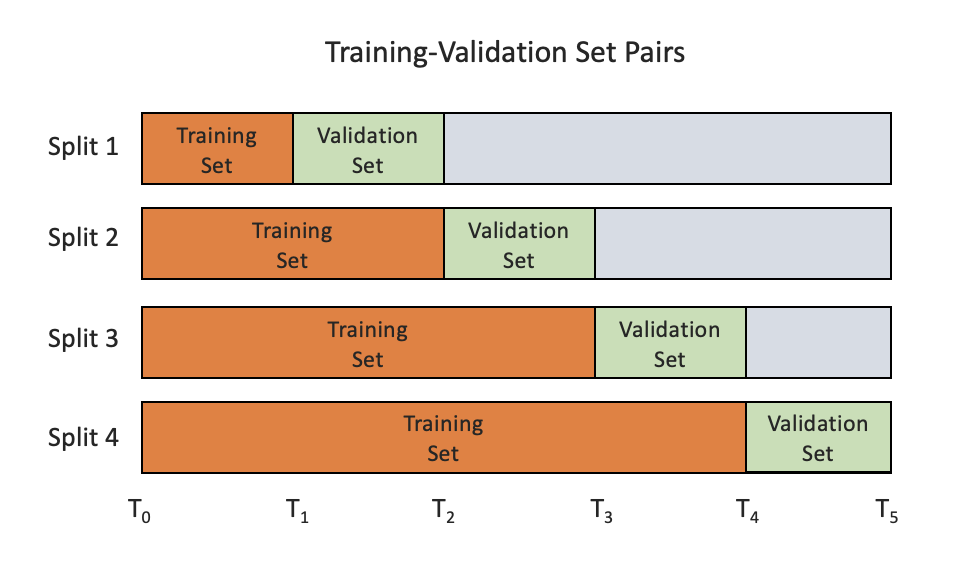}
    \caption{The temporal validation approach used in these settings to capture the non-stationary nature of the data and guard against leakage. Time is used to split the available data into a series of training and validation sets, testing for  generalization performance on ``future'' data relative to model training.}
    \label{fig:temporal_validation}
\end{figure}

\item{\textbf{Models}: We train a wide variety of model and hyperparameter combinations, including logistic regression, tree-based models, and ensembles such as random forests and boosted trees. The reasoning behind a wide grid was both to understand the effectiveness of different models along both the ``accuracy'' and bias dimensions as well as to provide the model selection process with as much diversity as possible. The model types and hyperparameters used for each problem are listed in Table~\ref{table:exp_details}}.

\item{\textbf{Choice of Bias Metric}: In all of these problems, a key decision to make is the choice of the appropriate bias metric(s). We use the Fairness Tree (Figure~\ref{fig:fairnesstree}), a framework developed and used in \cite{rodolfa2020case} to inform that choice. Since in all the problems we describe below, we are supporting assistive interventions (i.e. reducing disparities in false negatives is more important than those in false positives), and have limited resources to intervene compared to the number of people that need support, the Fairness Tree framework leads us to choose Recall (True Positive Rate) Disparity as the primary bias metric.}

\begin{figure*}
    \centering
    \includegraphics[width=1.0\linewidth]{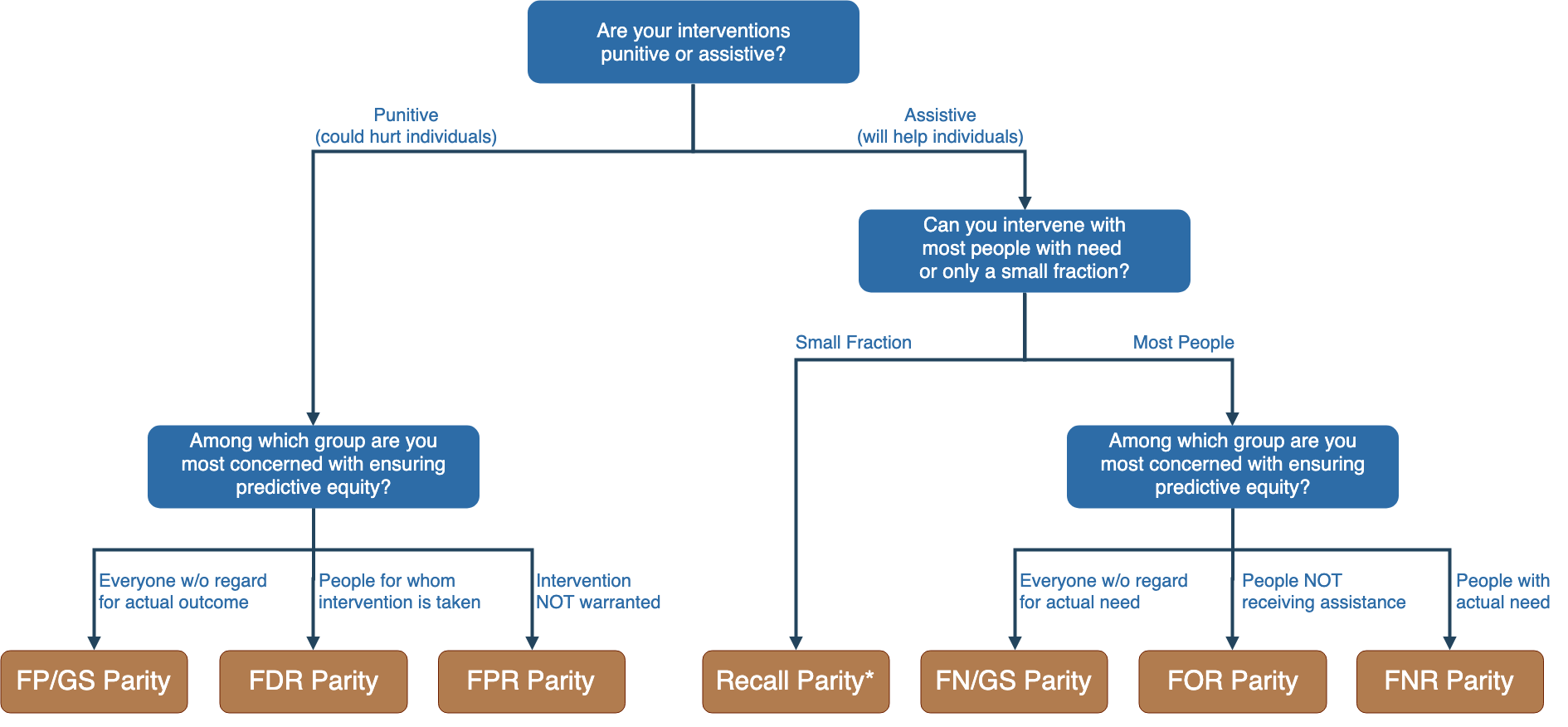}
    \caption{Fairness Tree framework to help identify appropriate fairness metrics based on the intended use. The metrics in the leaf nodes are: False Negative Rate (FNR), False Omission Rate (FOR), False Negatives Adjusted to Group Size (FN/GS), Recall/True Positive Rate (TPR), False Positive Rate (FPR), False Discovery Rate (FDR), and False Positives Adjusted to Group Size (FP/GS).}
    \label{fig:fairnesstree}
\end{figure*}

\item{\textbf{Evaluation Methodology}: For each temporal validation set we calculate the evaluation metric as well as the bias metric (with respect to the protected group) for all models. These results are aggregated by calculating the mean and standard errors.}

%\item{\textbf{Results}: In all the figures shown in this paper, we summarize the results by taking mean of the model's performance in validation set across different temporal blocks. The error estimates are shown by $95\%$ confidence intervals.}

\end{enumerate}

Each of the four policy problems used for the present empirical evaluation are described in detail below, including details about the underlying data set, the performance and fairness metrics of interest, and the protected group for bias and fairness analysis.

\begin{table*}[tbp]
\centering
\caption{Data and Experimental Setup for our four problems.}
\resizebox{\textwidth}{!}{%
\normalsize
\begin{tabular}{|l|c|c|c|c|}
\hline
 %& \multicolumn{4}{l|}{\textbf{Problem and Data}} \\ \hline
 & \multicolumn{1}{c|}{\textbf{\begin{tabular}[c]{@{}c@{}}Mental Health\\ and Criminal Justice\end{tabular}}} & \multicolumn{1}{c|}{\textbf{\begin{tabular}[c]{@{}c@{}}Housing Safety\\ Inspections\end{tabular}}} & \multicolumn{1}{c|}{\textbf{\begin{tabular}[c]{@{}c@{}}Student \\ Outcomes\end{tabular}}} & 
 \multicolumn{1}{c|}{\textbf{\begin{tabular}[c]{@{}c@{}}Education \\ Crowdfunding\end{tabular}}} \\ \hline
Prediction Task & \begin{tabular}[c]{@{}l@{}}Jail booking \\within the next \\12 months\end{tabular} & \begin{tabular}[c]{@{}l@{}}Housing unit having \\ a violation within \\ the next year\end{tabular} & \begin{tabular}[c]{@{}l@{}}Student not \\ returning to \\ school next year\end{tabular} & \begin{tabular}[c]{@{}l@{}}Project not getting \\ fully funded within \\ 4 months\end{tabular} \\ \hline
Timespan & 2013-01-01 to 2019-04-01 & 2011-01-01 to 2017-06-01 & 2009-01-01 to 2018-01-01 &  2010-01-01 to 2014-01-01 \\ \hline
\# of entities & 61,192 & 4,593 & 801,242 & 210,310 \\ \hline
Feature Groups 
    & 
    \specialcell{
        Demographics    \\
        Mental Health History   \\
        Past Diagnosis  \\
        Mental Health Programs  \\
        Police Interactions   \\
        Past Jail Incarceration \\
        Jail Booking Details
    }
    &  
    \specialcell{
        Building Permits    \\
        Past Citations  \\
        Past Violations \\
        House Prices    \\
        Census Data
    }
    &  
    \specialcell{
        Age Relative to Grade \\
        Repeated Grades    \\
        Rural/Urban \\
        Academic History   \\
        Dropout History   \\
        Gender  \\
        Illness \\
        Family Information
    }
    &  
    \specialcell{
        Funding Request Details \\
        Donation Details    \\
        Past Funding Rates  \\
        Project Description
    }
    \\ \hline
\# of Features & 3,465 & 1,657 & 220 & 319  \\ \hline
Base Rate & 0.12 & 0.43 & 0.25 & 0.24  \\ \hline
Evaluation Metric & Precision@500 & Precision@500 & Precision@10000 & Precision@1000  \\ \hline
\begin{tabular}[c]{@{}l@{}}Model Types and \\ Hyperparameters \\ (as specified by \\scikitlearn \\parameters \\used in the\\ experiments)\end{tabular} 
&
    \specialcell{
        \texttt{
            Decision Tree
        }   \\
        Max Depth: \\ (1,2,3)  \\
        Min Samples Split: \\ (10, 50, 100)\\
        \\
        \texttt{
            Random Forest
        }\\
        Num Estimators:\\ (100, 1000, 5000)    \\
        Min Samples Split:\\ (10, 25, 100) \\
        Max Depth: \\ (5, 10, 50)  \\
        \\
        \texttt{
            Logistic Regression
        }   \\
        Penalty: \\ (l1, l2)   \\
        C: \\ (0.001, 0.01, 0.1, 1, 10)
    }
&
    \footnotesize{\specialcell{
        \texttt{
            Decision Tree
        }   \\
        Criteria: \\ (gini, entropy)   \\
        Max Depth:  \\ (1,2,3,5,10,20,50)   \\
        Min Samples Split: \\ (10, 20, 50, 1000)\\
        \\
        \texttt{
            Random Forest
        }   \\
        Max Features: \\ (sqrt, log2)  \\
        Criteria: \\ (gini, entropy)   \\
        Num Estimators: \\ (100, 1000, 5000)   \\
        Min Samples Split: \\ (10, 20, 50, 100)    \\
        Max Depth: 
        \\
        (2, 5, 10, 20, 50, 100)  \\
        \\
        \texttt{
            Extra Trees
        }   \\
        Max Features: \\ (sqrt, log2)  \\
        Criterion: \\ (gini, entropy)  \\
        Num Estimators: \\ (100, 1000, 5000)   \\
        Min Samples Split: \\ (10, 20, 50, 100)    \\
        Max Depth: \\ (2, 5, 10, 50, 100)  \\
        \\
        \texttt{
            Logistic Regression
        }   \\
        Penalty: \\ (l1, l2)   \\
        C: \\ (0.001, 0.01, 0.1, 1, 10)   \\
        \\
       
    }}
&
    \specialcell{
        \texttt{Decision Tree}  \\
        Max Depth: \\ (1, 5, 10, 20, 50, 100)  \\
        Min Samples Split: \\ (2, 5, 10, 100, 1000)   \\
        \\
        \texttt{Extra Trees}    \\
        Num Estimators: (100)   \\
        Max Depth: (5, 50)  \\
        \\
        \texttt{Logistic Regression}    \\
        Penalty: \\ (l1, l2)   \\
        C: \\ (0.0001, 0.001, 0.1, 1, 10) \\
        
        \\
        \texttt{Random Forest}  \\
        Num Estimators: (100, 500)  \\
        Min Samples Split: (2, 10)  \\
         Class Weight: \\ (Balanced Subsample, Balanced) 
        \\
        Max Depth: (5, 50)
    }
&  
    \specialcell{
        \texttt{Random Forest}\\
        Num Estimators: \\ (100, 500, 1000)    \\
        Min Samples Split: \\ (10, 50) \\
        Max Depth: \\ (10, 50, 100)    \\
        \\
        \texttt{AdaBoost}   \\
        Num Estimators: (500, 1000) \\
        \texttt{Decision Tree}  \\
        Max Depth: \\ (1, 5, 10, 20, 50, 100)  \\
        Min Samples Split: \\ (2, 5, 10, 100, 1000)   \\
        \\
        \texttt{Logisitic Regression}   \\
        C: \\(0.0001, 0.001, 0.01, 0.1, 1, 10)    \\
        penalty: \\ (l1, l2)
    }
\\ \hline
\begin{tabular}[c]{@{}l@{}}Train and \\ Validation Sets\end{tabular} 
&
    \specialcell{
        Temporal Block:\\ 4 months
    }
&
    \specialcell{
        Temporal Block:\\ 2 months
    }
&
    \specialcell{
        Temporal Block:\\ 1 year
    }
&  
    \specialcell{
        Temporal Block:\\ 3 months
    }
\\ \hline
Protected Group & Race & Median Income & Age Relative to Grade & Poverty Level  \\ \hline
\end{tabular}%
}
\label{table:exp_details}
\end{table*}

\subsubsection{Mental Health Outreach - Johnson County KS}

Untreated mental health conditions often result in a negative spiral, which can culminate in repeated periods of incarceration with long term consequences both for the affected individual and the community as a whole~\cite{hamilton2010people}. Surveys of inmate populations have suggested a high prevalence of multiple and complex needs, with 64\% of people in local jails suffering from mental health issues and 55\% meeting criteria for substance abuse or dependence ~\cite{James2006MentalInmates}. The criminal justice system is poorly suited to address these needs, yet houses three times as many individuals with serious mental illness as hospitals~\cite{FullerTorrey2010MoreStates}.

Since 2016, Johnson County, KS, has partnered with our group to help them break this cycle of incarceration by identifying individuals who might benefit from outreach with mental health resources and are at risk for future incarceration. While the Johnson County Mental Health Center (JCMHC) currently provides services to the jail population, needs are generally identified reactively, for instance through screening instruments individuals fill out when entering jail. The new program being developed will supplement these existing approaches by adding a new automatic referral system for people who are at risk of being booked into jail, with the hope that they can be outreached to reduce their risk of returning to jail. 

Through our partnership, we obtained administrative data from their mental health center, jail system, police arrests, and ambulance runs. ML modeling was focused on Johnson County residents with any history of mental health need who had been released from jail within the past three years. Early results from this work were described in~\cite{bauman2018reducing}. A field evaluation of the predictive model is ongoing at the time of this writing, but validation on historical data demonstrated a 12\% improvement over a baseline based on the number of bookings in the prior year and 4.8-fold increase over the population prevalence.

\subsubsection{Housing Safety Inspections - San Jose, CA}
The Multiple Housing team in San Jose's Code Enforcement Office is tasked with protecting the occupants of properties with three or more units, such as apartment buildings, fraternities, sororities, and hotels. They do so by conducting routine inspections of these properties, looking for everything from blight and pest infestations to faulty construction and fire hazards (see \cite{holtzen2016perceptions} and \cite{klein2015affordable} for a discussion of the importance of housing inspections to public health). Although the city of San Jose inspects all of the properties on its Multiple Housing roster over time, and expects to find minor violations at many of them, it is important that they can identify and mitigate dangerous situations early to prevent accidents. With more than 4,500 multiple housing properties in San Jose, CA -- many of which comprise multiple buildings and hundreds of units -- it is not possible for the city to inspect every unit every year. San Jose recently instituted a tiered approach to prioritizing inspections, inspecting riskier properties more frequently and thoroughly. Although the tier system helped focus inspections on riskier properties, the new system has its limitations. The city evaluates tier assignments for properties infrequently (every 3 to 6 years), and these adjustments require a great deal of expertise and manual work while leaving out a rich amount of information. 

In order to provide a more nuanced view of properties' violation risk over time and allow for more efficient scheduling of inspections, the Code Enforcement Office partnered with us to develop a model to predict the risk that a serious violation would be found if a given property was prioritized for inspection (similar tools have been developed for allocating fire inspections in New York \cite{athey2017beyond} and health inspections in Boston \cite{glaeser2016crowdsourcing}). Evaluation of the model on historical data indicated that it could provide a 30\% increase in precision relative to the current tier system and the model's predictive accuracy was confirmed during a 4-month field trial in 2017.

\subsubsection{Improving Educational Outcomes - El Salvador}
Each year from 2010 through 2016, 15-29\% of students enrolled in school in El Salvador did not return to school in the following year. This high dropout rate is cause for serious concern, with significant consequences for economic productivity, workforce skill, inclusiveness of growth, social cohesion, and increasing youth risks \cite{belfield2007price,atwell2019building}. El Salvador's Ministry of Education has programs available to support students with the goal of reducing these high dropout rates, but the budget for these programs is not large enough to reach every student and school in El Salvador. 

Predictive modeling has been deployed to help schools identify students at risk of dropping out in several contexts \cite{lakkaraju2015machine,aguiar2015who,bowers2012dropout} and El Salvador partnered with us in 2018 to make use of these methods to focus their limited resources on the students at highest risk of not returning each year. Student-level data was provided by the Ministry of Education, including demographics, urbanicity, school-level resources (e.g., classrooms, computers, etc), gang and drug violence, family characteristics, attendance records, and grade repetition. For the present study, we focused on the state of San Salvador and identifying the 10,000 highest-risk students, considering annual cohorts of approximately 300,000 students and drawing on 5 years' of prior examples as training data.

\subsubsection{Education Crowdfunding - DonorsChoose} 
Since the projects above used confidential and sensitive data and were done under data use agreements, we are not able to make that data publicly available. For our work to be easily reproducible, we include a fourth problem in this study where the data is available publicly, focused around crowdfunding for education by the organization DonorsChoose.

Many schools in the United States, particularly in poorer communities, face funding shortages \cite{morgan2018funding}. Often, teachers themselves are left to fill this gap, purchasing supplies for their classrooms when they have the individual resources to do so \cite{huzra2015teachers}. The non-profit DonorsChoose was founded in 2000 to help alleviate these shortages by providing a platform where teachers post project requests focused on their classroom needs and community members can make  contributions to support these projects. Since 2000, they have facilitated \$970 million in donations to 40 million students in the United States \cite{DonorsChooseAbout}. However, approximately one third of all projects posted fail to reach their funding goal. 

Here, we make use of a dataset DonorsChoose made publicly available for the 2014 KDD Cup (an annual data science competition) including information about projects, the schools posting them, and donations they received. Because the other case studies explored here focused on proprietary and often sensitive data shared with us under data use agreements that cannot be made publicly available, we included a case study surrounding this publicly-available dataset. While we have not partnered with DonorsChoose to deploy the machine learning system described, we otherwise treated this case study as we would any of our applied projects. Here, we consider a resource-constrained effort to assist projects at risk of going unfunded (for instance, providing a review and consultation) capable of helping 1,000 projects in a 2-month window, focusing on the most recent 2 years' of data available in the extract (earlier data had far fewer projects and instability in the baseline funding rates as the platform ramped up). This dataset is publicly available at kaggle.com \cite{DonorsChooseData}.

\section{Results}
\begin{figure*}[!hbtp]
  \centering
  \begin{subfigure}[b]{0.99\textwidth}
    \centering
    \raisebox{0.1mm}{\includegraphics[width=\textwidth]{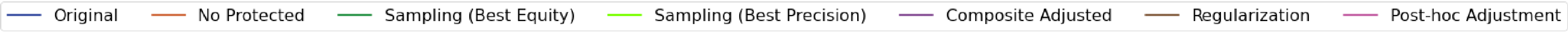}}
  \end{subfigure}
   \begin{subfigure}[b]{0.24\textwidth}
    \centering
    \raisebox{1mm}{\includegraphics[width=\textwidth]{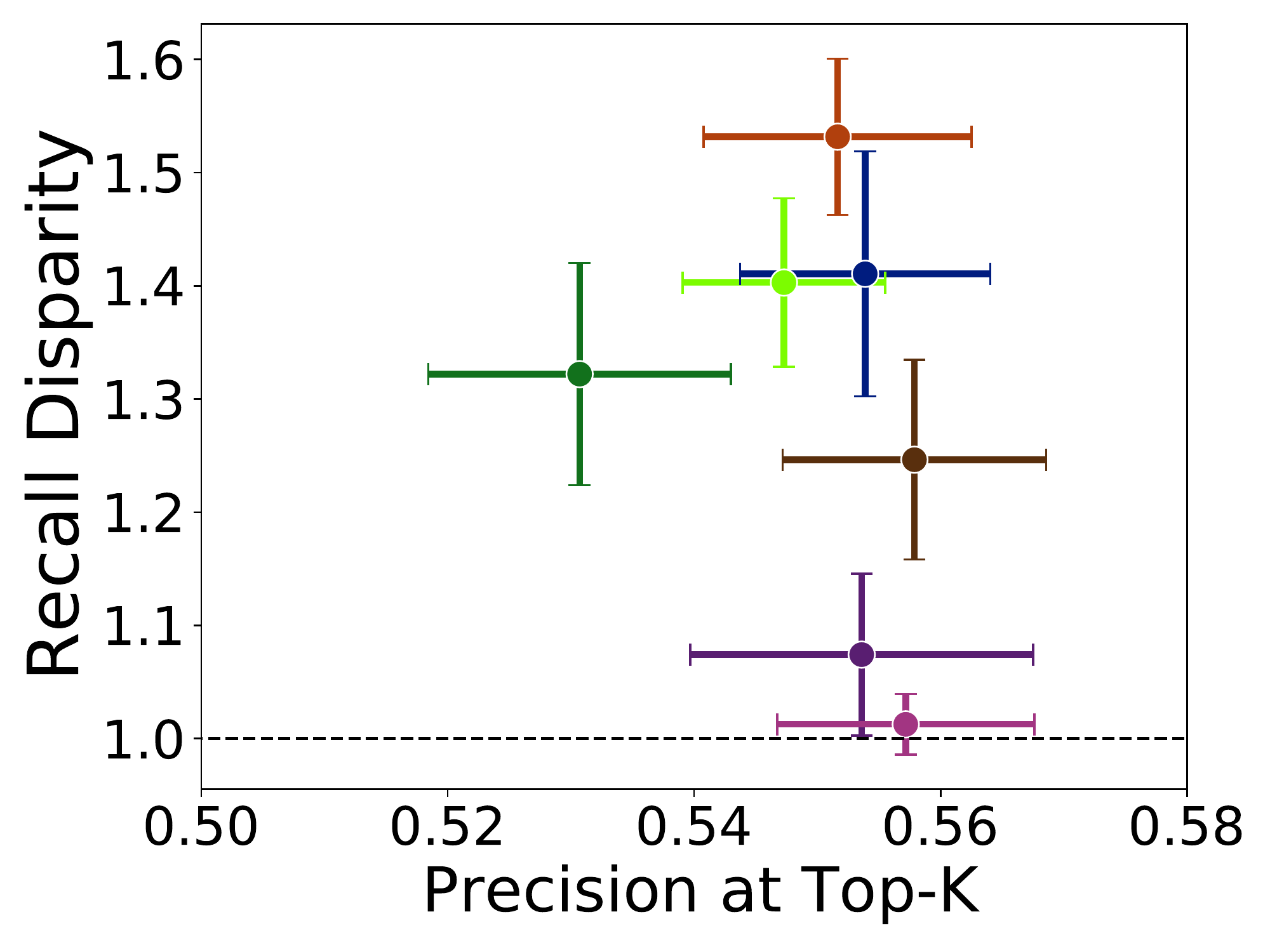}}%
    \caption{Inmate Mental Health}
   \end{subfigure}
   \begin{subfigure}[b]{0.24\textwidth}
    \centering
    \raisebox{1mm}{\includegraphics[width=\textwidth]{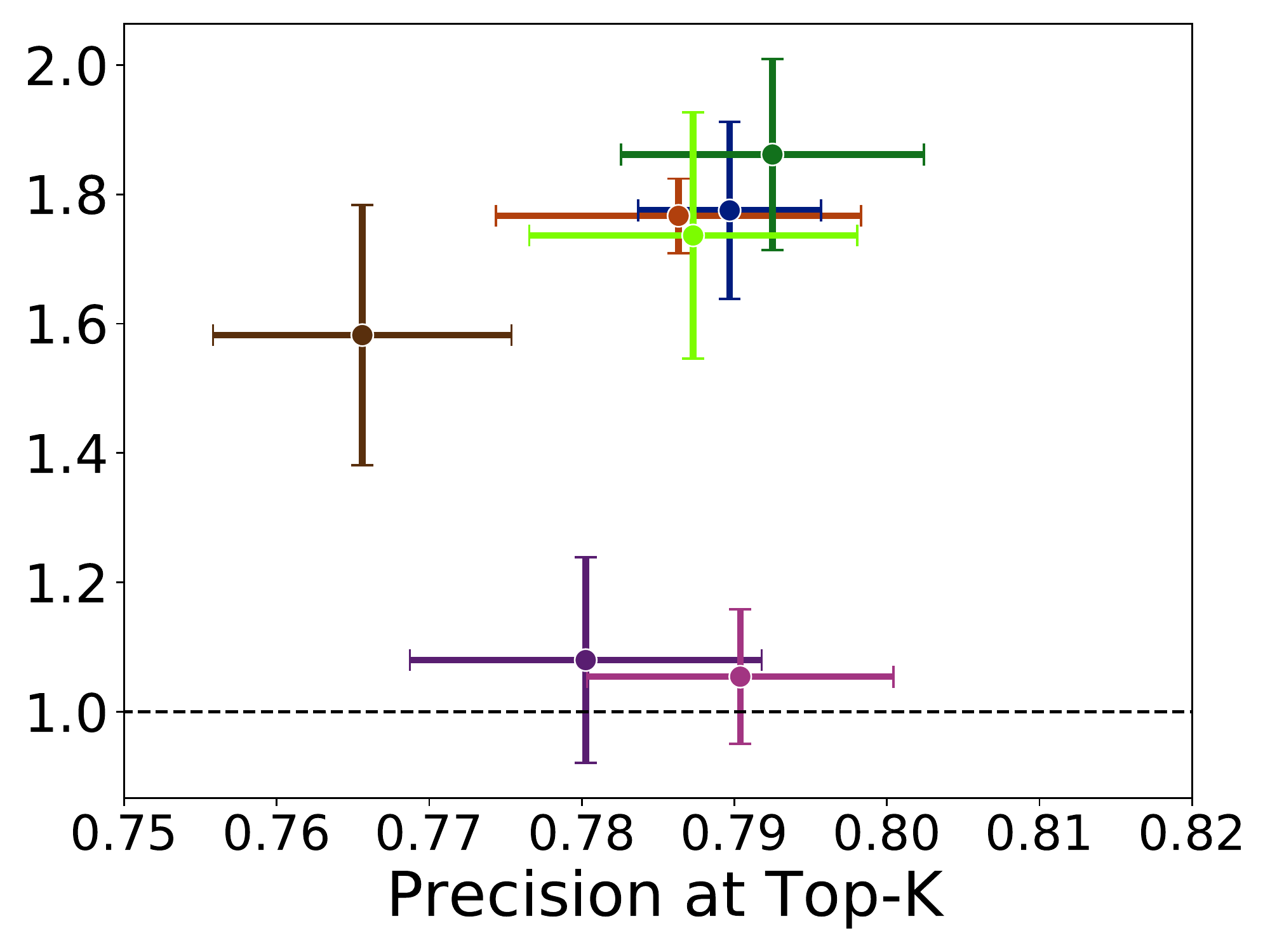}}%
    \caption{Housing Safety}
   \end{subfigure}
   \begin{subfigure}[b]{0.24\textwidth}
    \centering
    \raisebox{1mm}{\includegraphics[width=\textwidth]{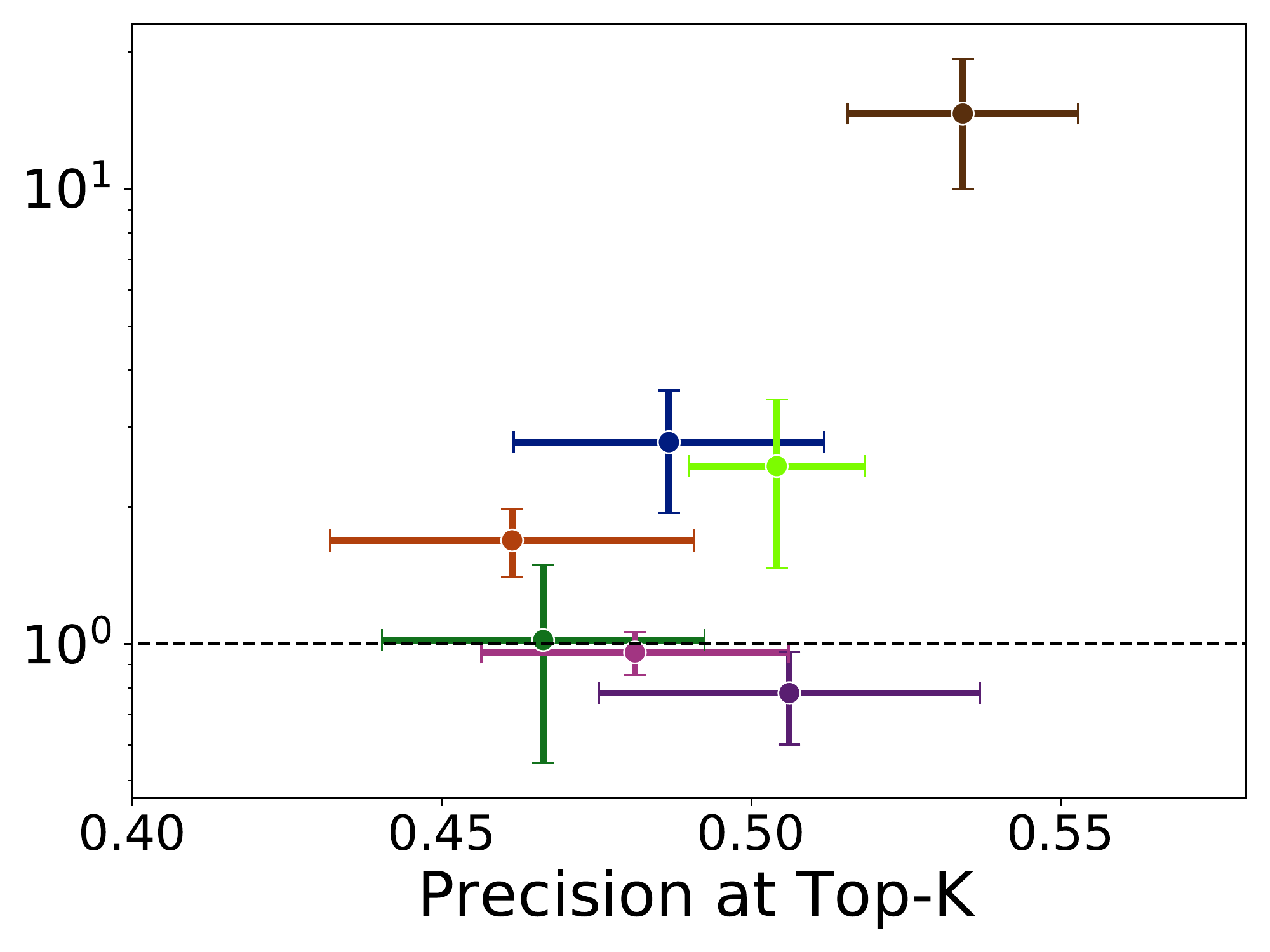}}%
    \caption{Student Outcomes}
   \end{subfigure}
   \begin{subfigure}[b]{0.24\textwidth}
    \centering
    \raisebox{1mm}{\includegraphics[width=\textwidth]{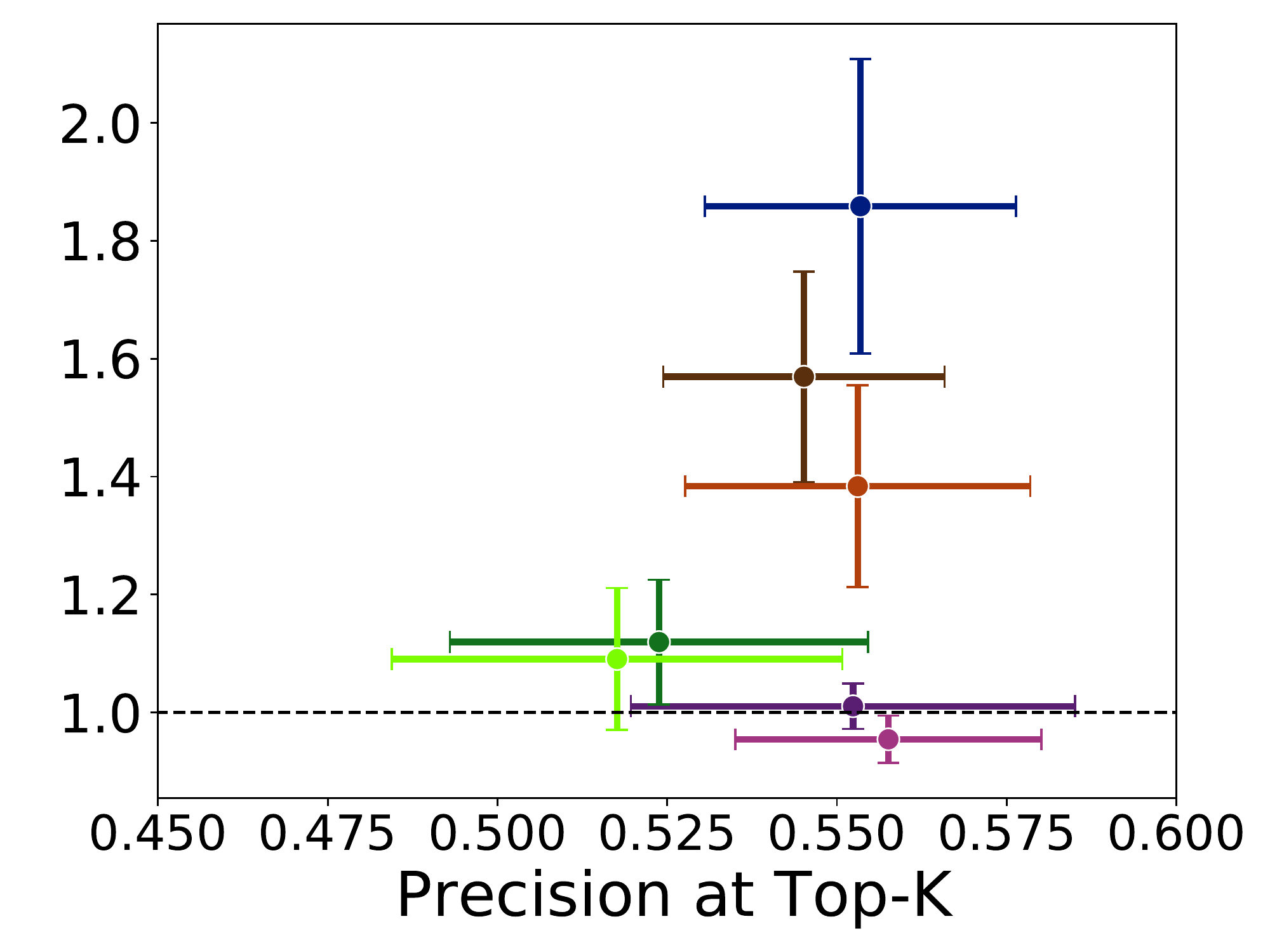}}%
    \caption{Education Crowdfunding}
   \end{subfigure}
   \caption{Results from the different fairness-enhancing strategies considered here across the four policy settings, showing the relationship between model accuracy (as measured by precision@k) on the x-axis and fairness (as measured by recall disparities) on the y-axis. Ideal models would have high values of precision@k and be near a disparity value of 1.0. Note that the y-axis in (c) is on a log scale based on the large variation in performance across methods (performance for each method is shown separately on a linear scale in the figures below). Error bars show 95\% confidence intervals over validation sets.}
   \label{fig:overall}
\end{figure*}

Overall results across the different methods and problems we evaluated are shown in Figure \ref{fig:overall}. Each graph shows the relative performance of models with a given strategy in terms of the ``performance metric'' (namely precision@k on the x-axis) and fairness with respect to the protected group (namely True Positive Rate or Recall disparities on the y-axis), with error bars representing the 95\% confidence interval over all temporal validation splits. The ideal model would have a value of 1.0 for both of these metrics -- models appearing further to the right on the x-axis are more accurate while those appearing closer to the dashed y=1.0 line are more equitable (departures from this line in either direction reflect disparities favoring one or the other group).

The blue circle in all of the graphs refers to the \textit{Original} model --- a term we use to specify the model built and selected only focused on maximizing the accuracy metric. All the other points are results from the bias reduction methods that we investigated. Note that in this figure we only include the best-performing sampling strategies (either in terms of fairness or accuracy) but discuss and show the wider range of sampling results in the graphs below. Likewise, we only show the composite models without decoupled training because the results from the two strategies were generally similar, but discuss and show these results in more detail below as well. Additionally, model selection approaches are omitted from Figure \ref{fig:overall} because they generally required considerable decreases in precision@k to improve fairness, allowing us to focus the overall analysis on the nuance between the other methods (see Figure \ref{fig:model_selection} and the related discussion for these results).

\subsection{Overall Results}

Across the four problems, a few general patterns seem to emerge from our experiments: 

\begin{enumerate}
    \item \textbf{Considerable disparities, ranging from 30-100\%, were observed in the \textit{baseline} models} for all four problems. That is, building models which optimize only for some measure of accuracy consistently resulted in appreciable biases if fairness was not actively pursued as an outcome. This is not a surprising outcome and a result that has been demonstrated in various studies but an important point to keep in mind when building ML models.  
    
    \item There was \textbf{considerable variability across strategies and settings} in the effectiveness and ability of the fairness-enhancing methods considered here to remove these disparities, with \textbf{most methods showing only moderate success} or success only in a few settings. Comparisons across these methods is discussed in more detail below. 
    
    \item Only the two approaches which made use of \textbf{separate thresholds across groups (composite models and post-hoc adjustments) were consistently successful} in removing disparities and did so without any appreciable loss in model accuracy. 
\end{enumerate}

Below we discuss these results in more detail, examining the performance of each fairness-enhancing approach in turn.

% Before presenting all the results in details, we first present a summary of the key takeaways:
% \begin{itemize}
% \item{All the non bias-mitigated models, which have been designed only to improve efficency are not equitable.}
% \item{Most of the bias reduction methods do reduce disparities to some extent but with high variance and varying costs in terms of efficiency.}
% \item{Post-hoc correction method was the only method that consistently resulted in reducing disparities without any significant loss in ``accuracy"}
% \end{itemize}

\subsection{Effect of Removing Sensitive Attribute}
A common misconception in the context of algorithmic fairness is that simply omitting a sensitive attribute can help a model achieve fair predictions through ``unawareness.'' Several authors~\cite{kleinberg2018algorithmic,pedreshi2008discrimination} have spoken to the fallacy of this concept, both as a result of correlations between protected attributes and other potentially relevant ones and because access to the sensitive attribute might help models pick up on patterns that improve accuracy for the protected group and result in lower disparities. However, we included this strategy here both for completeness as well as to understand and demonstrate how this approach might perform in practice. 

\begin{figure*}[!hbtp]
  \centering
   \begin{subfigure}[b]{0.24\textwidth}
    \centering
    \raisebox{1mm}{\includegraphics[width=\textwidth]{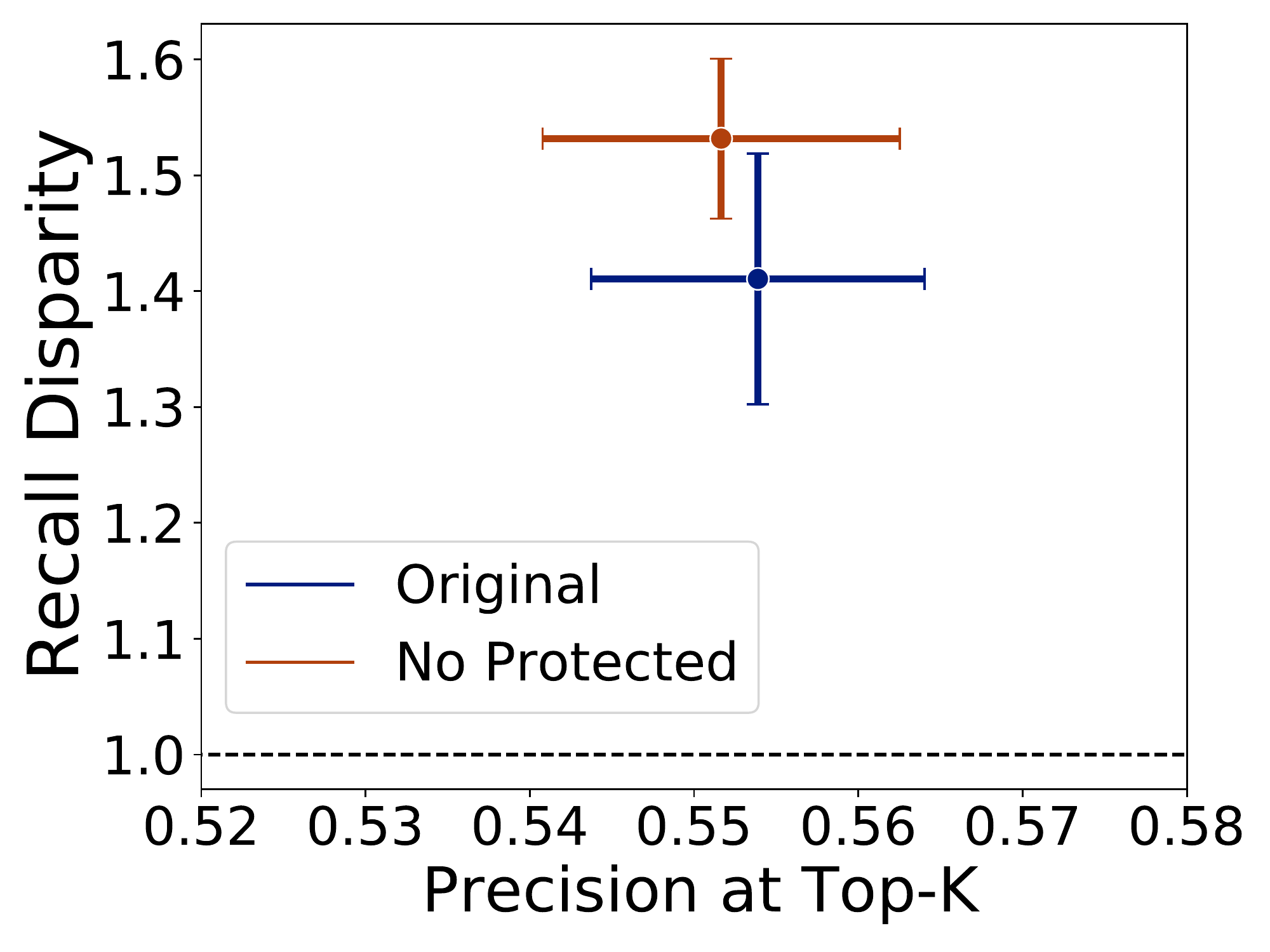}}%
    \caption{Inmate Mental Health}
   \end{subfigure}
   \begin{subfigure}[b]{0.24\textwidth}
    \centering
    \raisebox{1mm}{\includegraphics[width=\textwidth]{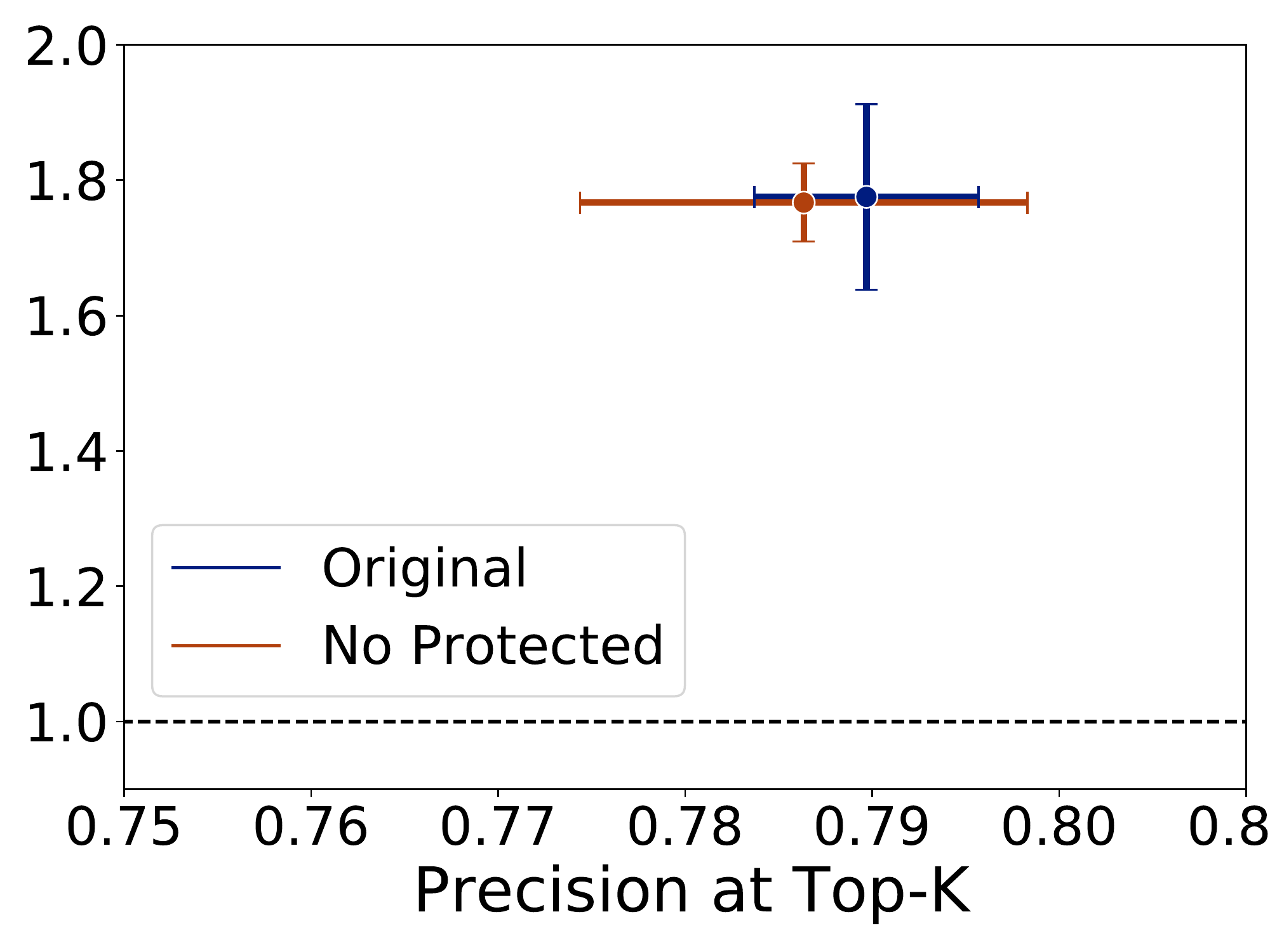}}%
    \caption{Housing Safety}
   \end{subfigure}
   \begin{subfigure}[b]{0.24\textwidth}
    \centering
    \raisebox{1mm}{\includegraphics[width=\textwidth]{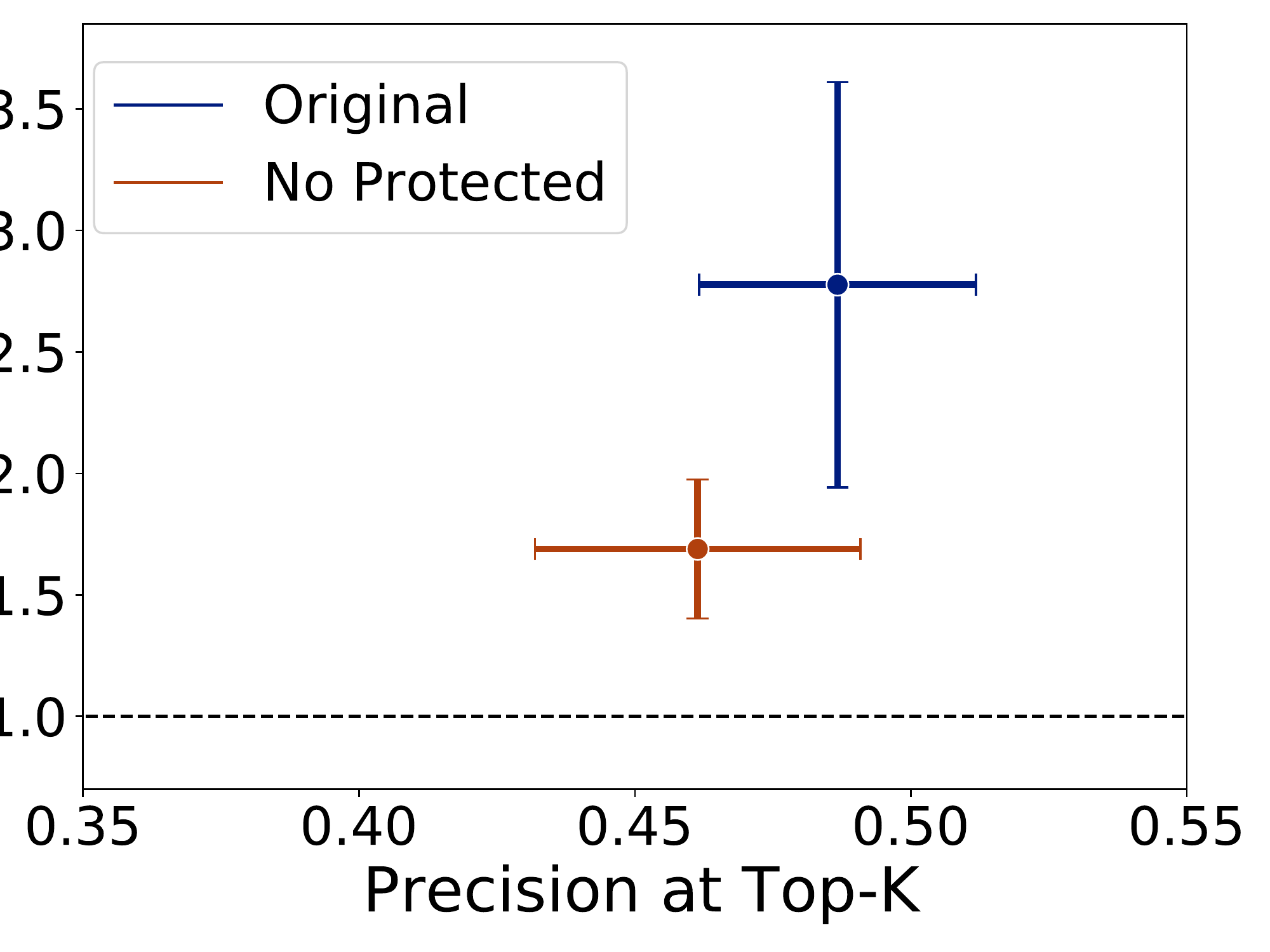}}%
    \caption{Student Outcomes}
   \end{subfigure}
   \begin{subfigure}[b]{0.24\textwidth}
    \centering
    \raisebox{1mm}{\includegraphics[width=\textwidth]{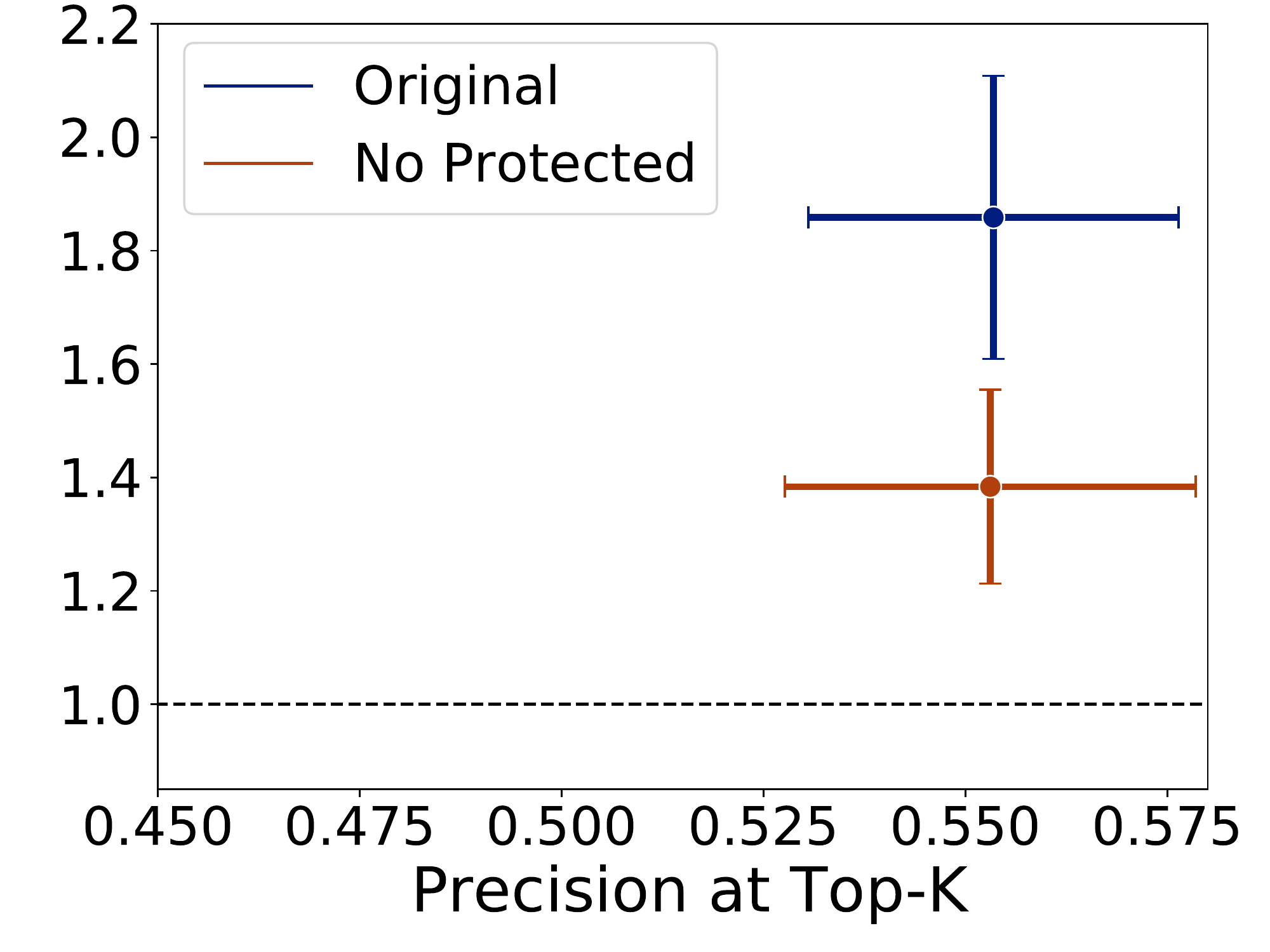}}%
    \caption{Education Crowdfunding}
   \end{subfigure}
   \caption{Effect of removing the protected attribute from machine learning modeling on model accuracy (precision@k) and fairness (recall disparities). Error bars show 95\% confidence intervals over validation sets.}
   \label{fig:no_protected}
\end{figure*}

Unsurprisingly, then, the results in Figure \ref{fig:no_protected} show \textbf{this strategy is inconsistent in the magnitude and direction of its impact across the four problems}. Although omitting the protected attribute did improve model fairness somewhat in the Education Crowdfunding and Student Outcomes contexts, in neither case did it fully remove the disparities from the initial model, and in the latter case these improvements came at the cost of a moderate decrease in precision@k. Moreover, in the Inmate Mental Health setting, removing the race attribute actually made the models somewhat less fair on average while in the Housing Safety context doing so had no effect on either fairness or accuracy. Taken together, these results are very consistent with the notion that ``fairness through unawareness'' by \textbf{removing the sensitive feature cannot be relied upon to improve the fairness of machine learning models}.

\subsection{Effect of Sampling}
The other pre-processing method we explored involved sampling of the training data. As discussed above, a number of parameters must be determined in choosing a sampling strategy: the relative distributions of the protected and non-protected subgroups, the label distributions within each group, and whether to over- or under-sample training examples to achieve the target distribution. Here, we explored six strategies (Table \ref{table:sampling}) that reflect combinations of three reasonable hypothesis:
\begin{itemize}
    \item A 1:1 ratio between protected group and non-protected group training examples might tell the model to treat errors in each group as equally important, alleviating differential error rates.
    \item Equal label distributions within the two subgroups might tell the model to not treat protected group membership as important.
    \item A 50/50 label distribution in one or both subgroup might alleviate any issues arising from imbalance in the training set.
\end{itemize}
%\krcomment{maybe better to move this discussion above where we introduce sampling?}

\begin{figure*}[!hbtp]
  \centering
   \begin{subfigure}[b]{0.24\textwidth}
    \centering
    \raisebox{1mm}{\includegraphics[width=\textwidth]{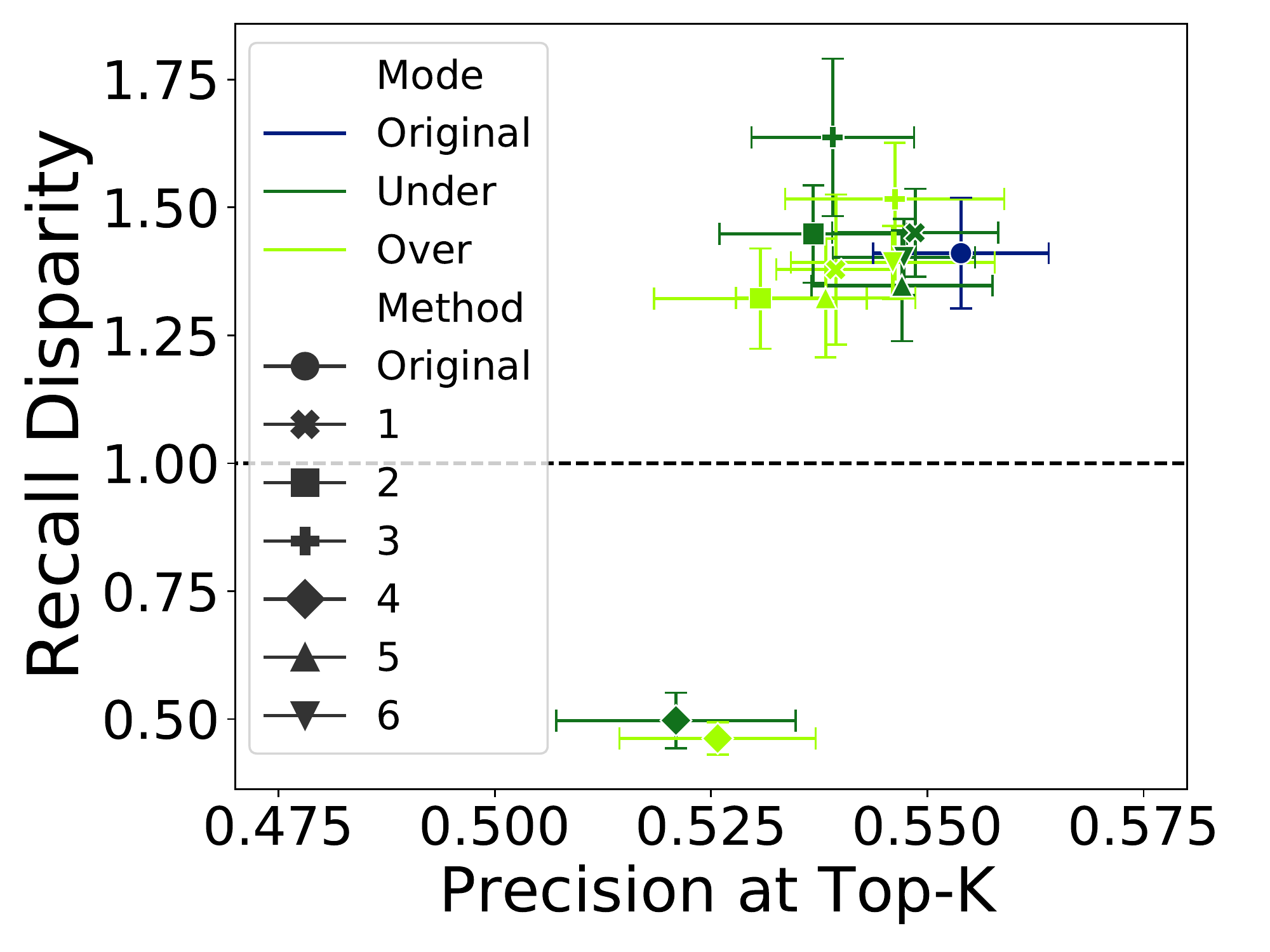}}%
    \caption{Inmate Mental Health}
   \end{subfigure}
   \begin{subfigure}[b]{0.24\textwidth}
    \centering
    \raisebox{1mm}{\includegraphics[width=\textwidth]{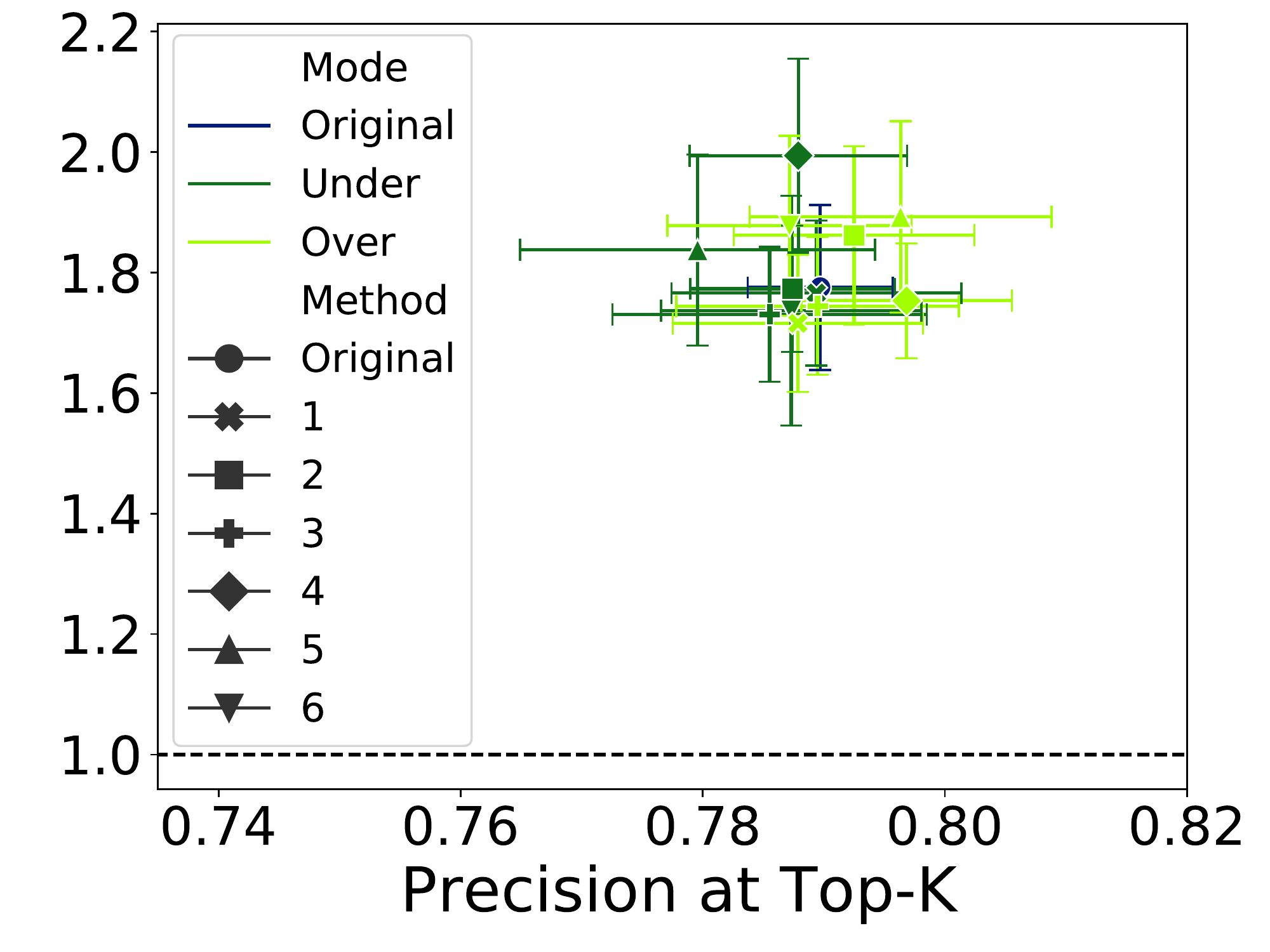}}%
    \caption{Housing Safety}
   \end{subfigure}
   \begin{subfigure}[b]{0.24\textwidth}
    \centering
    \raisebox{1mm}{\includegraphics[width=\textwidth]{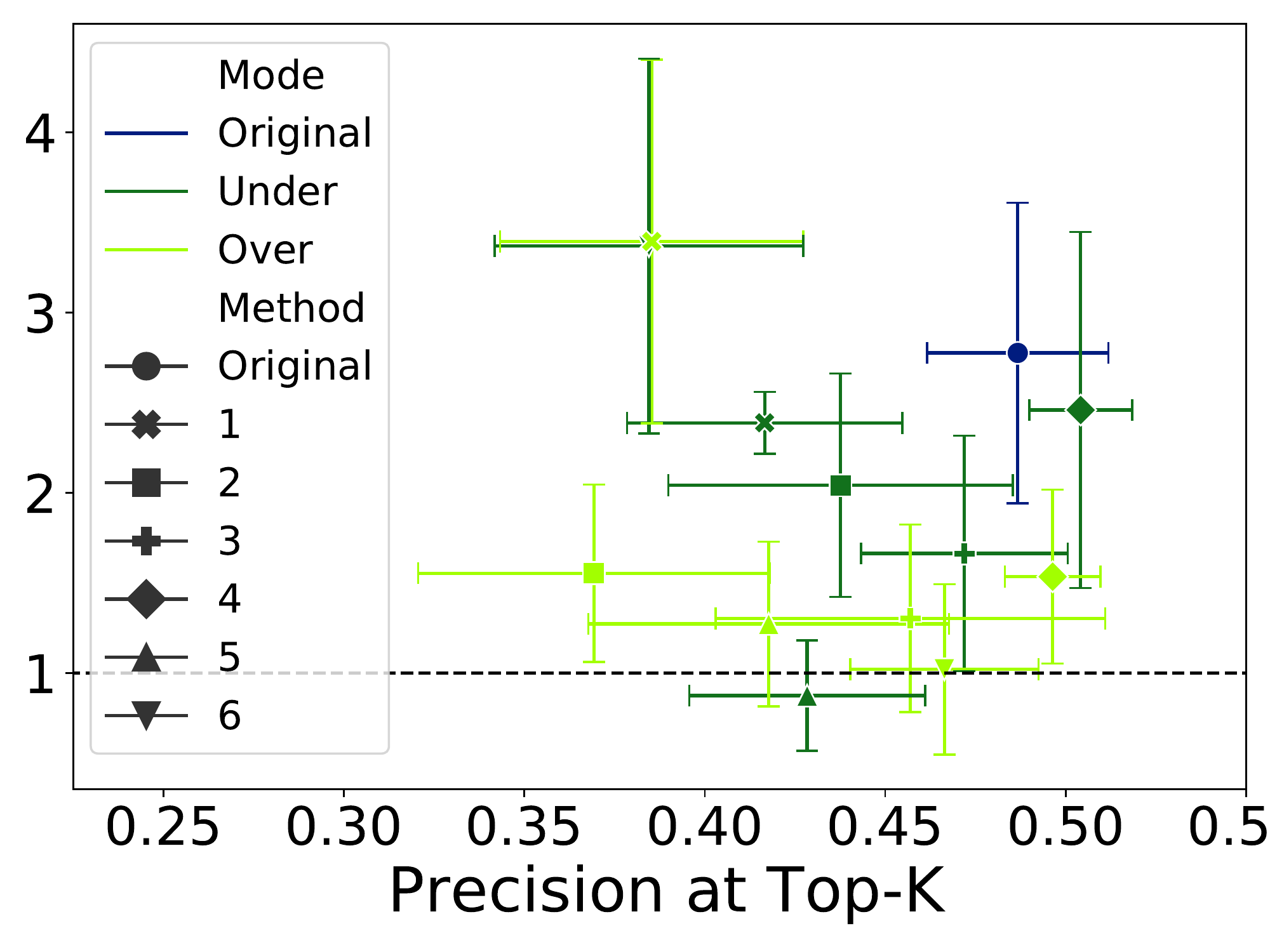}}%
    \caption{Student Outcomes}
   \end{subfigure}
   \begin{subfigure}[b]{0.24\textwidth}
    \centering
    \raisebox{1mm}{\includegraphics[width=\textwidth]{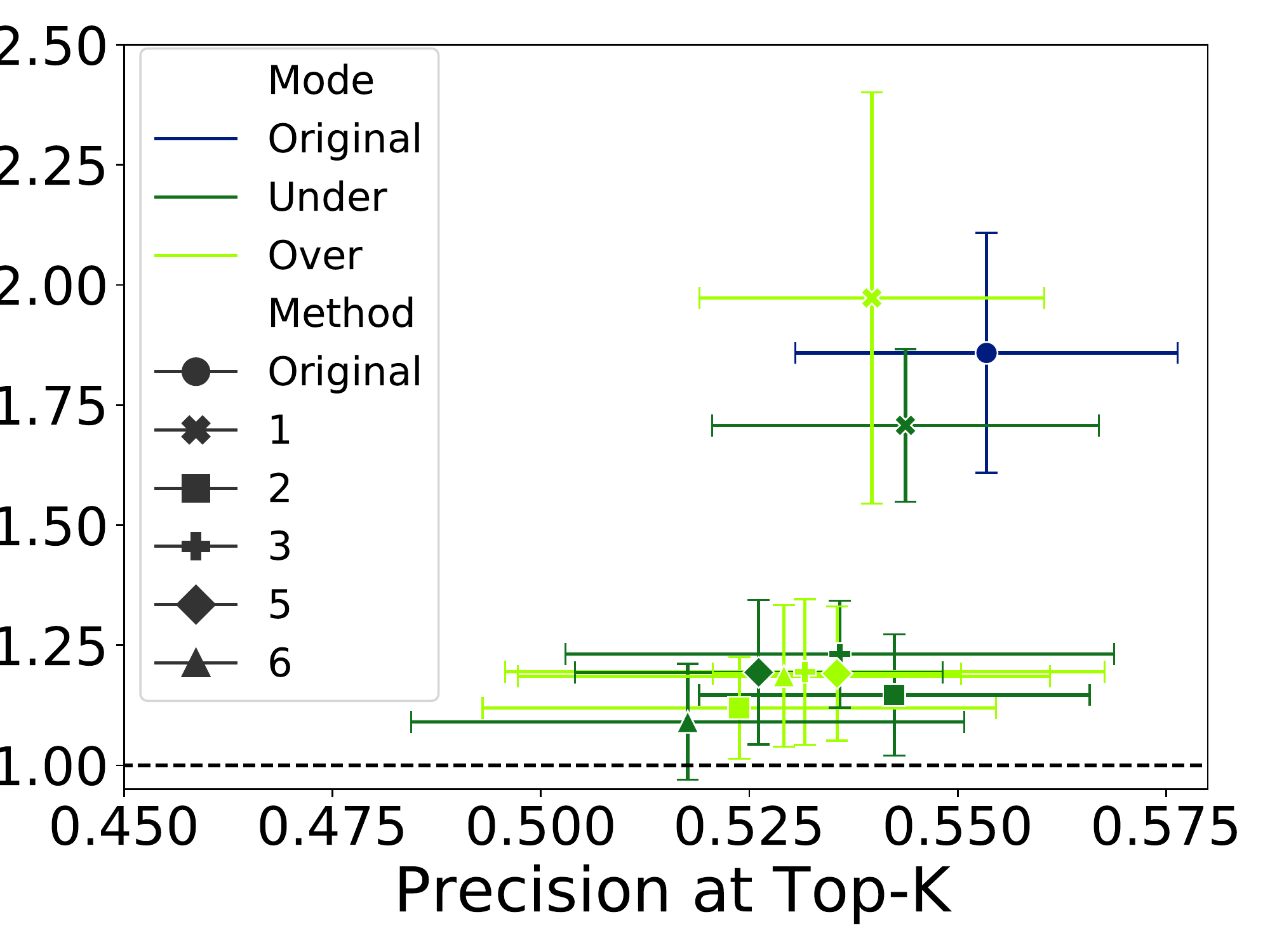}}%
    \caption{Education Crowdfunding}
   \end{subfigure}
   \caption{Results from resampling of training data for machine learning on model accuracy (precision@k) and fairness (recall disparities). Each of the six strategies from Table \ref{table:sampling} was performed either via under-sampling (dark green) or over-sampling (light green). Error bars show 95\% confidence intervals over validation sets.}
   \label{fig:sampling}
\end{figure*}

Figure \ref{fig:sampling} shows the results of applying these six strategies to the training data in each of the four policy settings. Although resampling of the training data had an impact on the models in many of the problem settings, there was a considerable inconsistency in the results both across settings and sampling strategies. In the Education Crowdfunding and Student Outcomes settings, many (but not all) of the strategies showed improvements in model fairness, while none of the strategies yielded fair results in the Housing Safety or Inmate Mental Health settings. Interestingly, this pattern reflects the results observed when removing the protected attribute described above, suggesting that both strategies may be accomplishing the same thing by effectively telling the model not to treat subgroup membership as important. Note, in particular, that in the Education Crowdfunding setting, sampling strategies 2, 3, 5, and 6 show improvements and each of these strategies equalizes the label distribution across the protected and non-protected subgroups.

Two more general patterns in Figure \ref{fig:sampling} do seem of note: First, over- and under-sampling approaches to the sample sampling strategy appear to yield similar results, suggesting that, at least in these four policy contexts, decreasing the total number of training examples through undersampling did not have an appreciable impact on model performance. Second, strategy 4 yielded particularly variable results, ranging from little impact to large disparities in either direction (note that in the Education Crowdfunding setting, both over- and under-sampling for strategy 4 resulted in no predicted positives in the protected class, yielding infinite disparities, so this strategy is omitted from the graph). However, this result might not be too surprising in light of the fact that this is the only strategy considered here where we adjusted the label distribution among the protected subgroup (to 50/50) without changing the distribution of non-protected subgroup. Depending on the baseline distribution, of course, this might (or might not) tell the model to see the protected attribute (or correlated features) as particularly important as a predictor of the outcome.

Taken together, these results suggest that \textbf{sampling of the training data can have an impact on disparities in the resulting models' predictions, but that these effects are both context and parameter dependent}. Without an obvious or consistent pattern for how a given sampling strategy will translate into changes in fairness metrics in a given modeling context, model developers are left to conduct a search over different values of these sampling parameters to explore this space in their setting. Even so, \textbf{there does not seem to be strong empirical evidence that a fairness-enhancing solution will be found in a particular context, or at what cost to model accuracy}.

\subsection{Effect of In-Processing Methods}
\begin{figure*}[!hbtp]
  \centering
   \begin{subfigure}[b]{0.24\textwidth}
    \centering
    \raisebox{1mm}{\includegraphics[width=\textwidth]{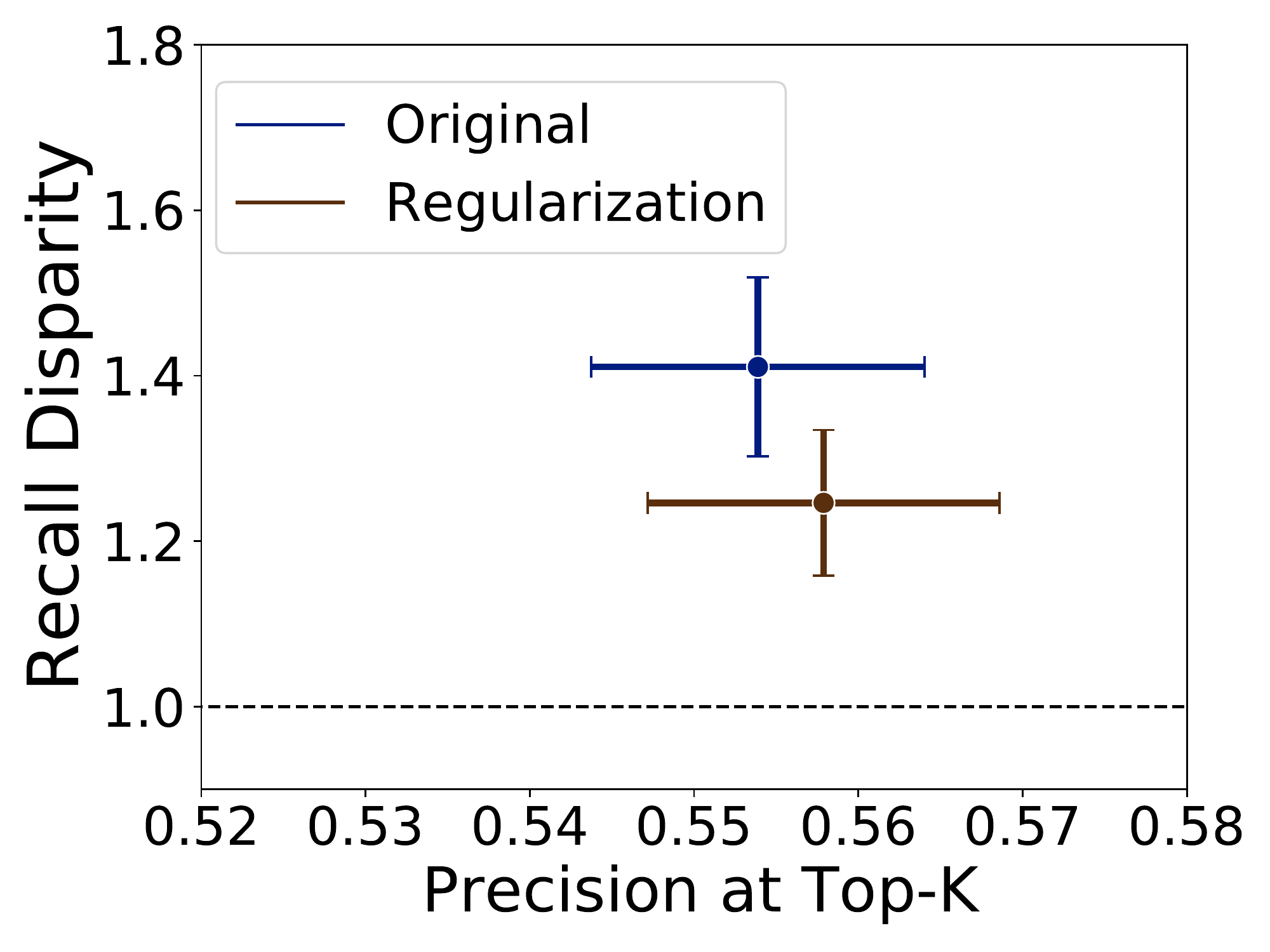}}%
    \caption{Inmate Mental Health}
   \end{subfigure}
   \begin{subfigure}[b]{0.24\textwidth}
    \centering
    \raisebox{1mm}{\includegraphics[width=\textwidth]{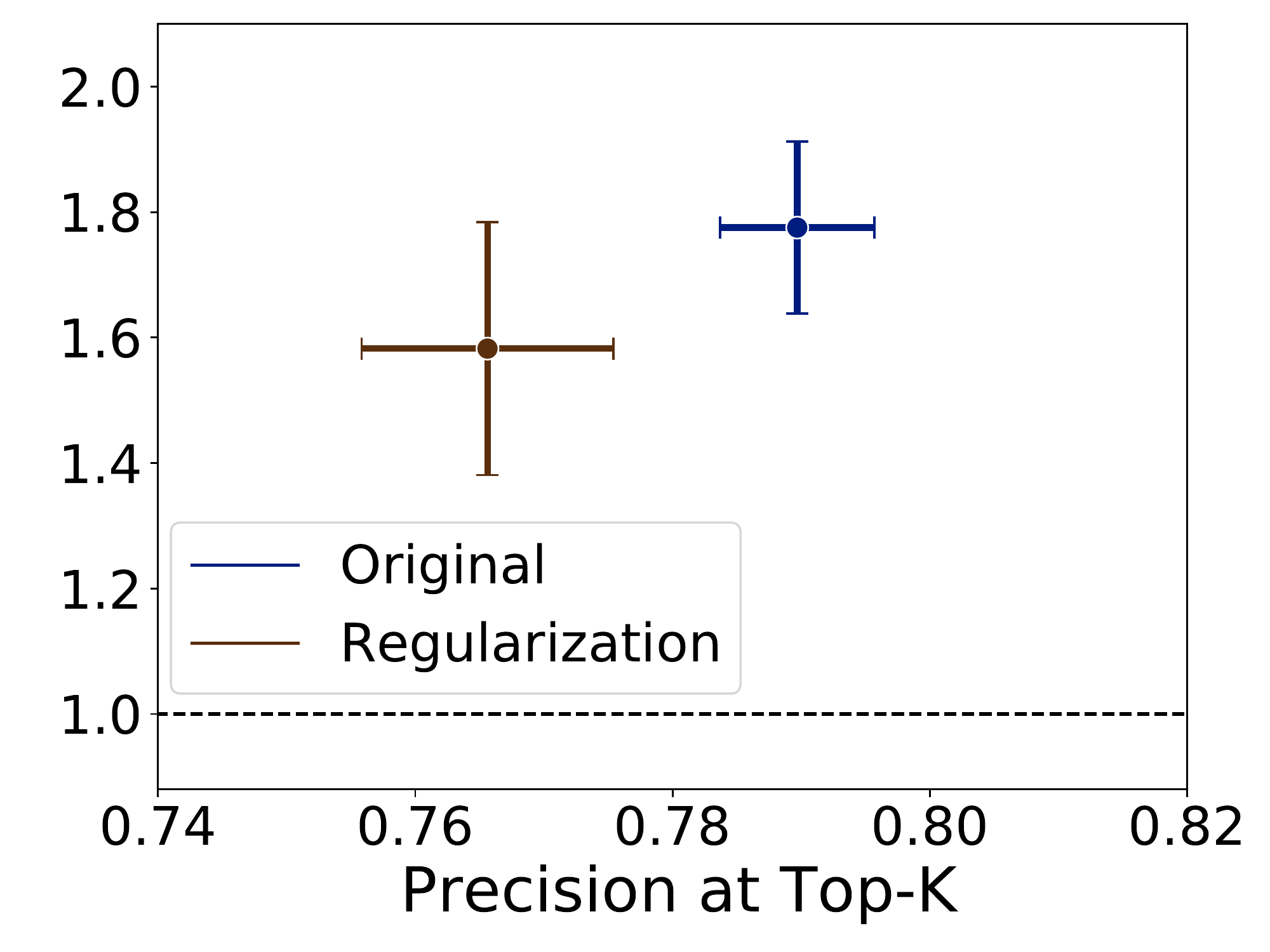}}%
    \caption{Housing Safety}
   \end{subfigure}
   \begin{subfigure}[b]{0.24\textwidth}
    \centering
    \raisebox{1mm}{\includegraphics[width=\textwidth]{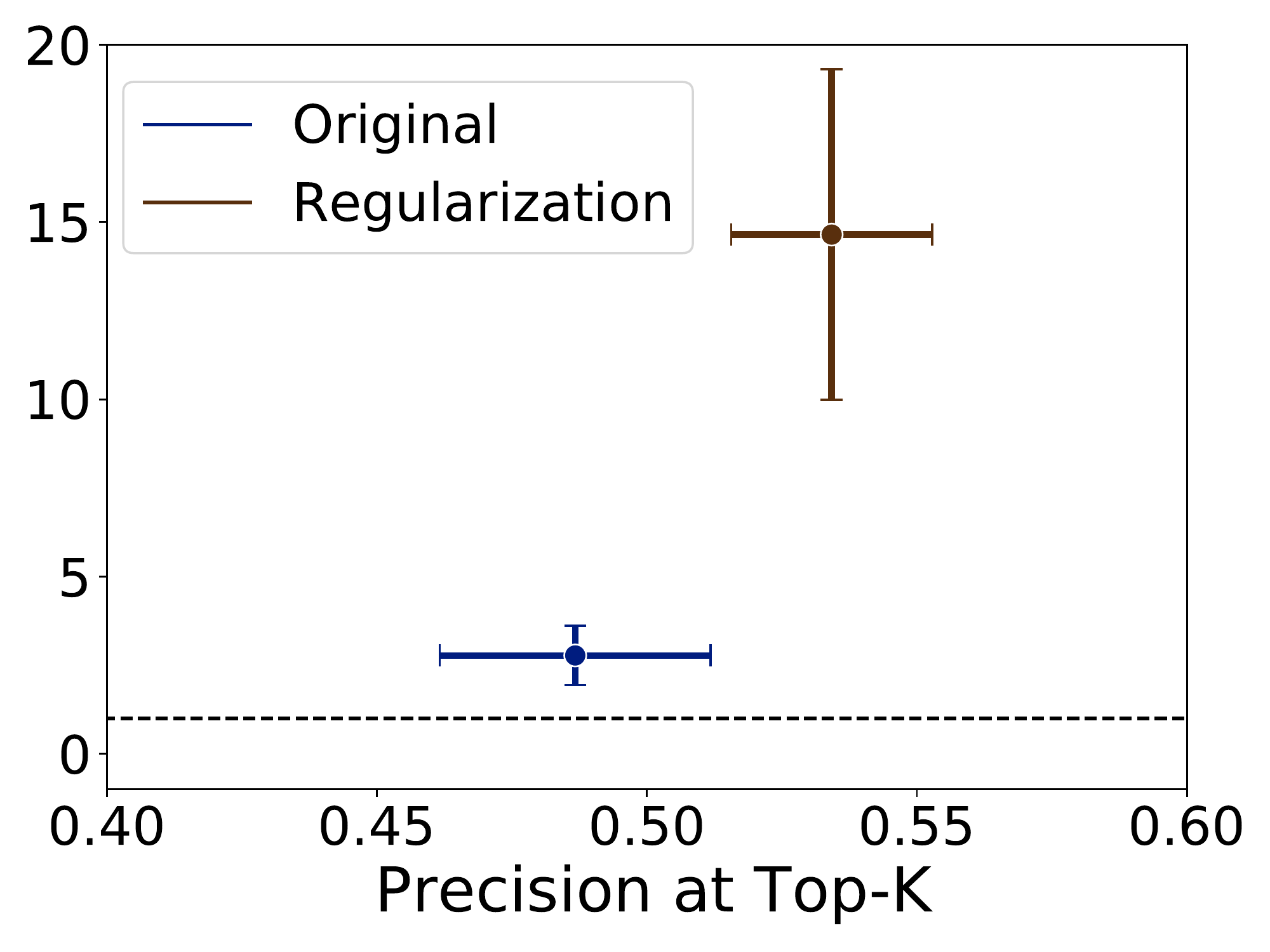}}%
    \caption{Student Outcomes}
   \end{subfigure}
   \begin{subfigure}[b]{0.24\textwidth}
    \centering
    \raisebox{1mm}{\includegraphics[width=\textwidth]{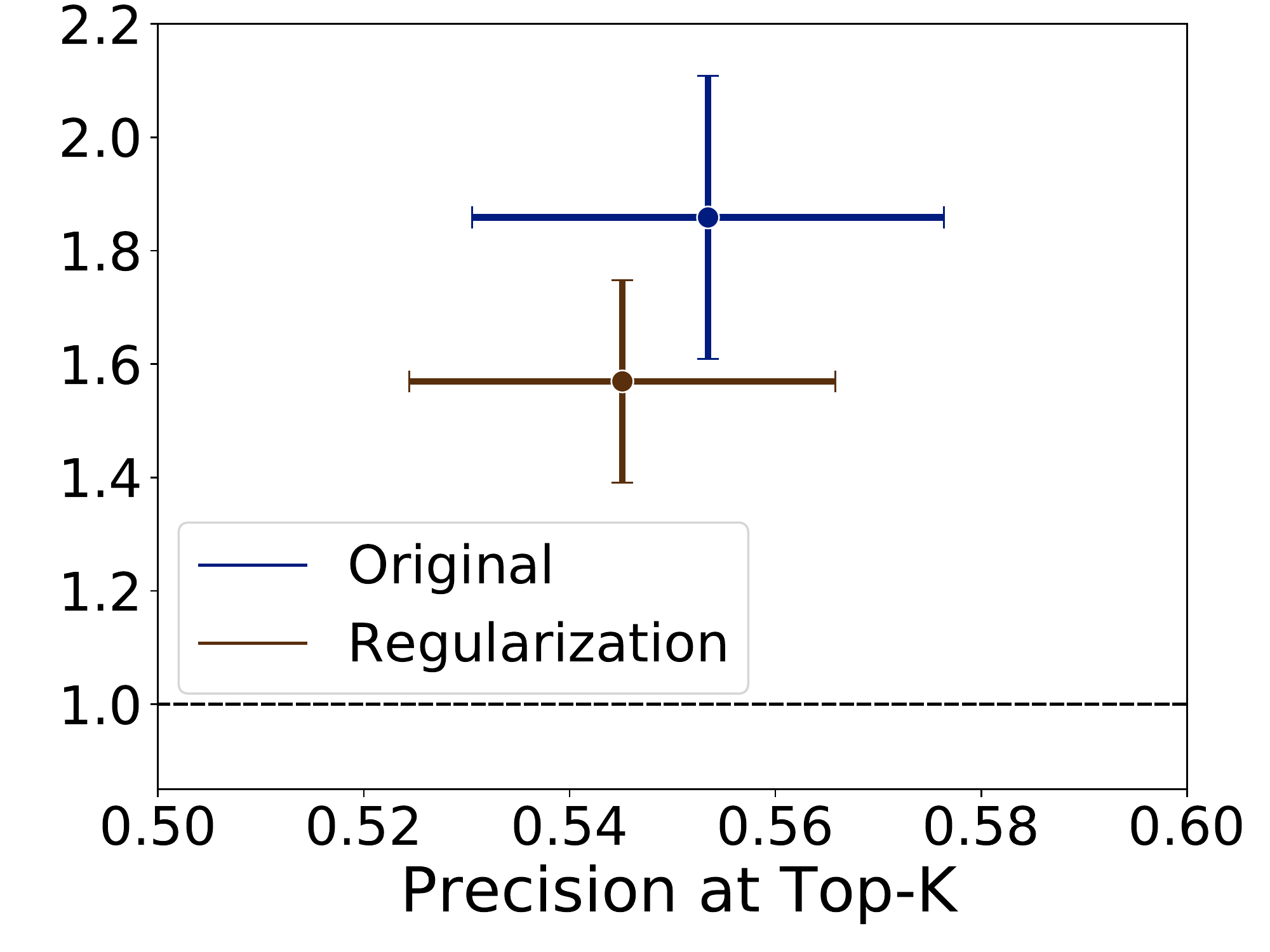}}%
    \caption{Education Crowdfunding}
   \end{subfigure}
   \caption{Results of using the in-processing method proposed by Zafar and colleagues to perform fairness-constrained optimization during model training on model accuracy (precision@k) and fairness (recall disparities). Error bars show 95\% confidence intervals over validation sets.}
   \label{fig:in_processing}
\end{figure*}

The in-processing method considered here was proposed by Zafar and colleagues~\cite{Zafar2017FairnessClassification}. Models were trained with a constraint to equalize the false negative rate\footnote{Note that $FNR = 1-TPR$, so this constraint is equivalent equalizing TPR across groups.} between the protected and non-protected group in each setting, with results shown in Figure~\ref{fig:in_processing}. In general, in-processing failed to appreciably improve the fairness of the models in any of the four settings, reducing disparities only slightly in three settings and making them appreciably worse in the fourth (Student Outcomes).

Importantly, these results should not be seen as an inherent critique of either Zafar's method specifically or in-processing in general, but rather as a mismatch between the currently available methods using this approach and the common context of resource-constrained problem settings in which a given number of highest-risk entities must be selected for an intervention. In-processing methods generally add a fairness constraint to a classifier that optimizes for overall accuracy around an implicit threshold of 0.5 (or best-partitioning decision boundary). To select the ``top k'' for intervention, we naively threshold the resulting score (or, equivalently, shift the decision boundary) to yield only $k$ highest-risk predicted positives. Of course, the fairness constraints used during model training applied to the original boundary, not the shifted one. As such, it is not surprising that Zafar's method here failed to improve fairness of these models when applied to a ``top k'' setting, even if it might perform well in settings without such a constraint. Perhaps for this reason, other methods such as Microsoft's Fair Learn~\cite{bird2020fairlearn} only provide predicted class labels without a continuous score, but unfortunately those methods are also poorly suited to the ``top k'' setting where a small subset of $k$ individuals would need to be randomly chosen from the predicted positive class at considerable cost to accuracy/precision.\footnote{We explored this package in particular in the Education Crowdfunding setting in a recent tutorial presented at the 2020 KDD and 2021 AAAI conferences~\cite{saleiro2020dealing}.}

\subsection{Effect of Model Selection}
\begin{figure*}[!hbtp]
  \centering
   \begin{subfigure}[b]{0.25\textwidth}
    \centering
    \raisebox{1mm}{\includegraphics[width=\textwidth]{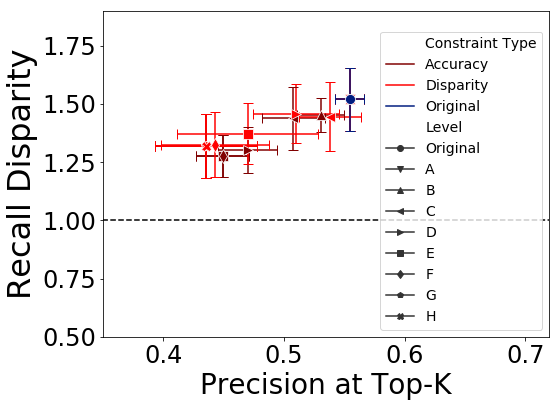}}%
    \caption{Inmate Mental Health}
   \end{subfigure}
   \begin{subfigure}[b]{0.25\textwidth}
    \centering
    \raisebox{1mm}{\includegraphics[width=\textwidth]{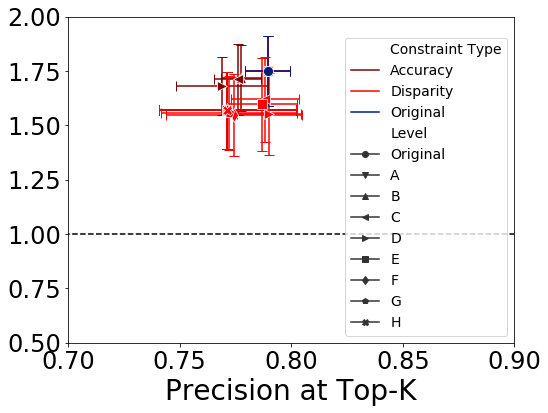}}%
    \caption{Housing Safety}
   \end{subfigure}
   \begin{subfigure}[b]{0.22\textwidth}
    \centering
    \raisebox{1mm}{\includegraphics[width=\textwidth]{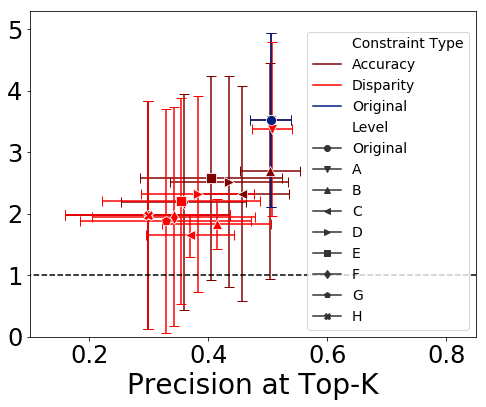}}%
    \caption{Student Outcomes}
   \end{subfigure}
   \begin{subfigure}[b]{0.24\textwidth}
    \centering
    \raisebox{1mm}{\includegraphics[width=\textwidth]{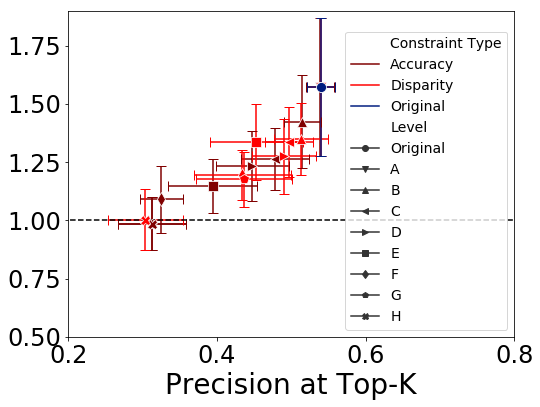}}%
    \caption{Education Crowdfunding}
   \end{subfigure}
   \caption{Effect of fairness-aware model selection on model accuracy (precision@k) and fairness (recall disparities). Model selection was performed either by setting a maximum acceptable disparity and choosing the model with the best precision@k among these (Disparity Constraint) or setting a maximum acceptable decrease in precision@k and choosing the lowest-disparity model among these (Accuracy Constraint). For each type, eight levels of constraint were explored (labeled A-H in the figure, from least to most weight on fairness). For Disparity Constraints, these are: A: 5.0, B: 2.0, C: 1.5, D: 1.3, E: 1.2, F: 1.1, G: 1.05, H: 1.0; for Accuracy Constraints, these are: A: 0.0, B: 0.05, C: 0.10, D: 0.15, E: 0.2, F: 0.25, G: 0.5, H: 0.6. Error bars show 95\% confidence intervals over validation sets.}
   \label{fig:model_selection}
\end{figure*}

Results of applying fairness-aware model selection in these contexts are shown in Figure \ref{fig:model_selection}. Here, several of the settings suggest an often considerable trade-off between fairness and accuracy, with constraints that put more weight on fairness in the model selection process yielding sizable decreases in precision@k (note that the range of the x-axes for these graphs is generally much wider than for the results of using other methods). For instance, in the Education Crowdfunding setting, disparities could be removed through the model selection process, but at the expense of loosing nearly half of the model's precision. In other cases, even large fairness constraints in the model selection process failed to remove disparities effectively: even when sacrificing a large amount of precision in the Inmate Mental Health context, the resulting models still showed considerable disparities of 1.27 on average. Likewise, in the Housing Safety context, fairness-aware model selection failed to reduce the disparities in these models regardless of constraint type or size. Notably, across all four contexts, similar results could be obtained by placing either a soft constraint on the largest acceptable disparity or a hard constraint on the largest acceptable decrease in precision@k to improve fairness (represented by different colors in Figure \ref{fig:model_selection}).

Although on the surface, these results suggest some semblance of the ``Pareto Frontier'' one might anticipate could reflect an inherent trade-off between fairness and accuracy, it is important to keep in mind that the nature of this frontier is highly dependent on the model grid over which this selection process is taking place (that is, other model type/hyperparameter combinations may perform better on one or both metrics). Likewise, other approaches at improving model fairness (such as the other methods explored here) may expand this frontier and allow for considerably less drastic trade-offs between fairness and accuracy.

\subsection{Effect of Post-Hoc Adjustments}
\begin{figure*}[!hbtp]
  \centering
   \begin{subfigure}[b]{0.24\textwidth}
    \centering
    \raisebox{1mm}{\includegraphics[width=\textwidth]{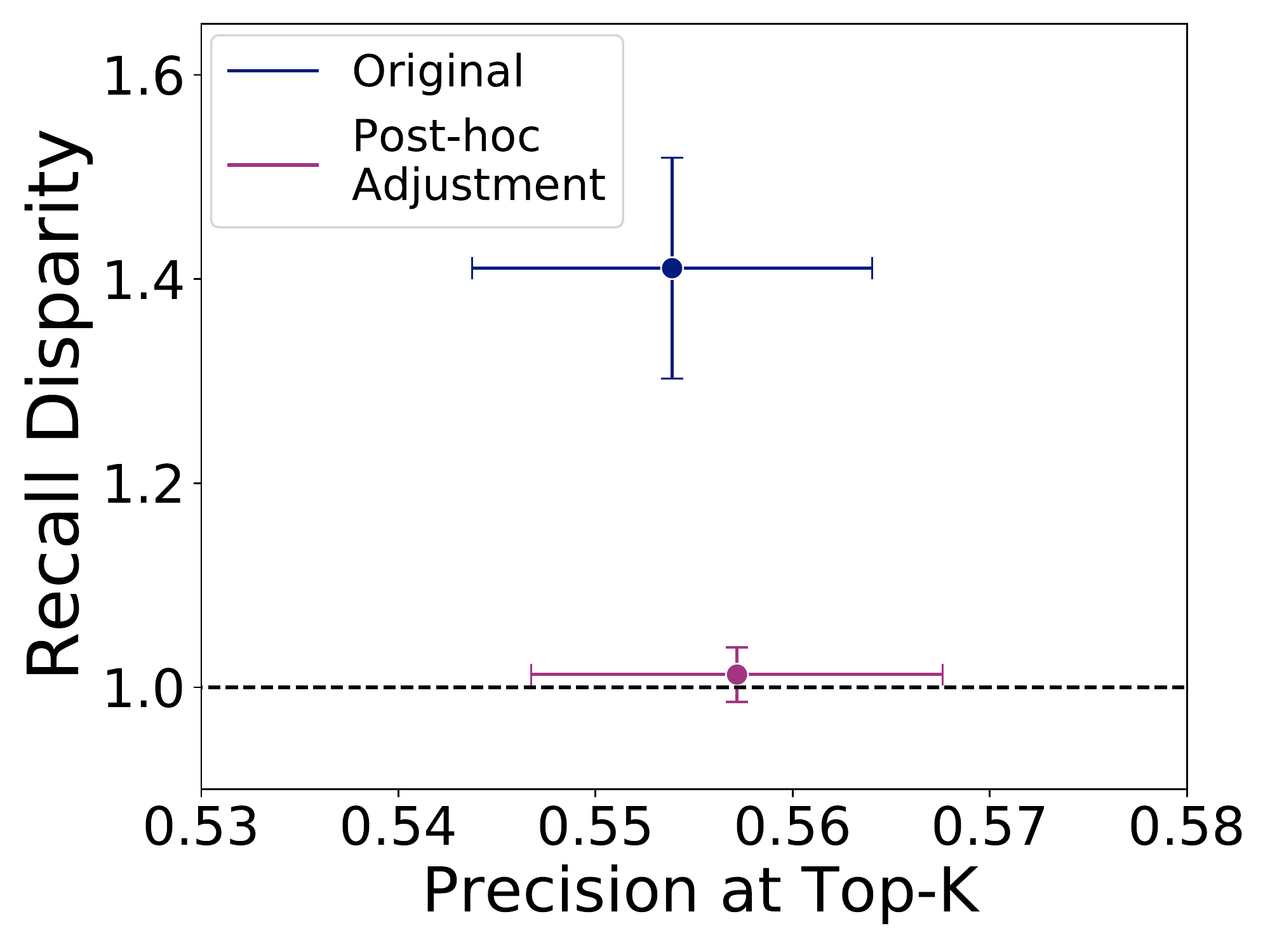}}%
    \caption{Inmate Mental Health}
   \end{subfigure}
   \begin{subfigure}[b]{0.24\textwidth}
    \centering
    \raisebox{1mm}{\includegraphics[width=\textwidth]{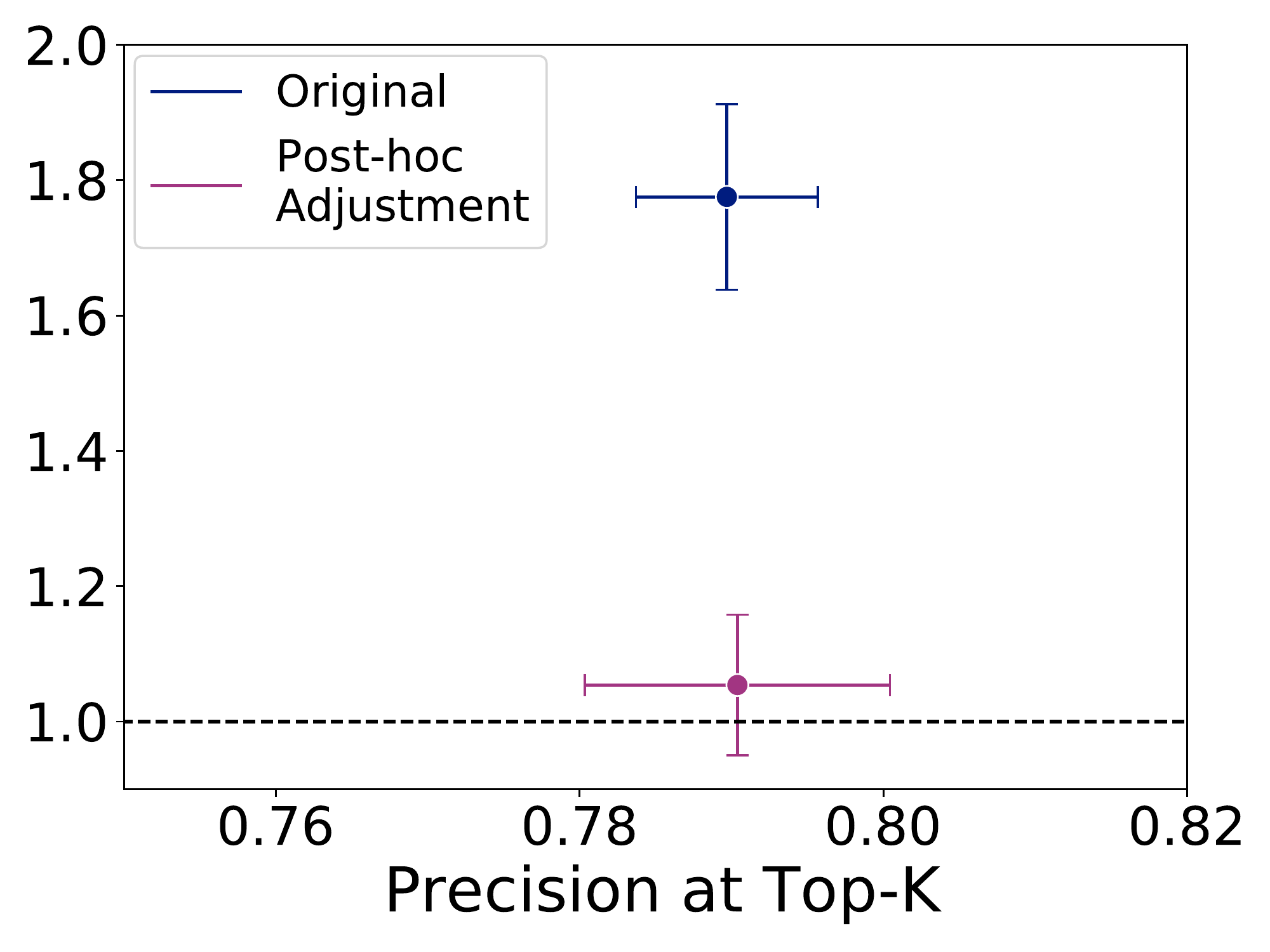}}%
    \caption{Housing Safety}
   \end{subfigure}
   \begin{subfigure}[b]{0.24\textwidth}
    \centering
    \raisebox{1mm}{\includegraphics[width=\textwidth]{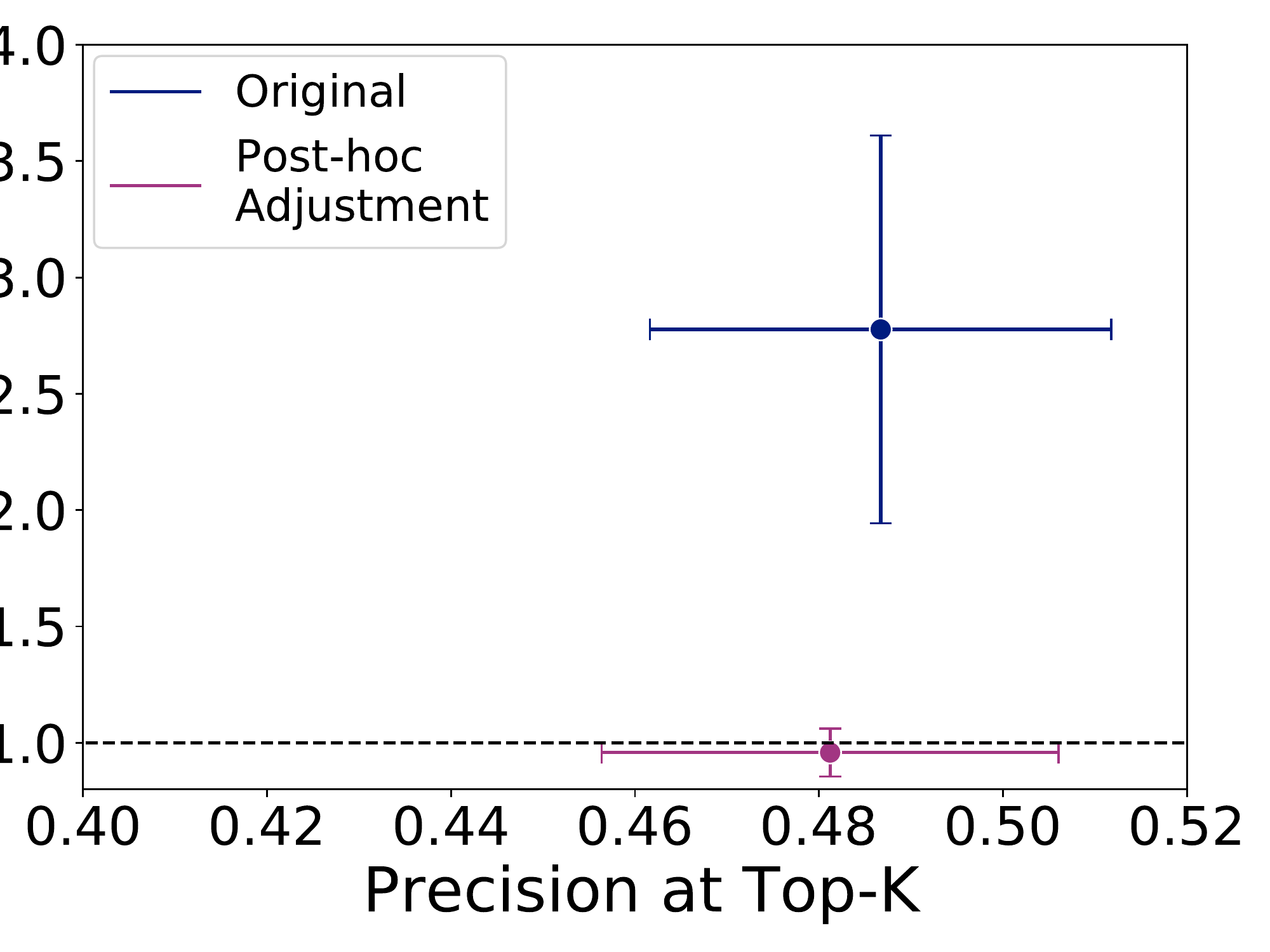}}%
    \caption{Student Outcomes}
   \end{subfigure}
   \begin{subfigure}[b]{0.24\textwidth}
    \centering
    \raisebox{1mm}{\includegraphics[width=\textwidth]{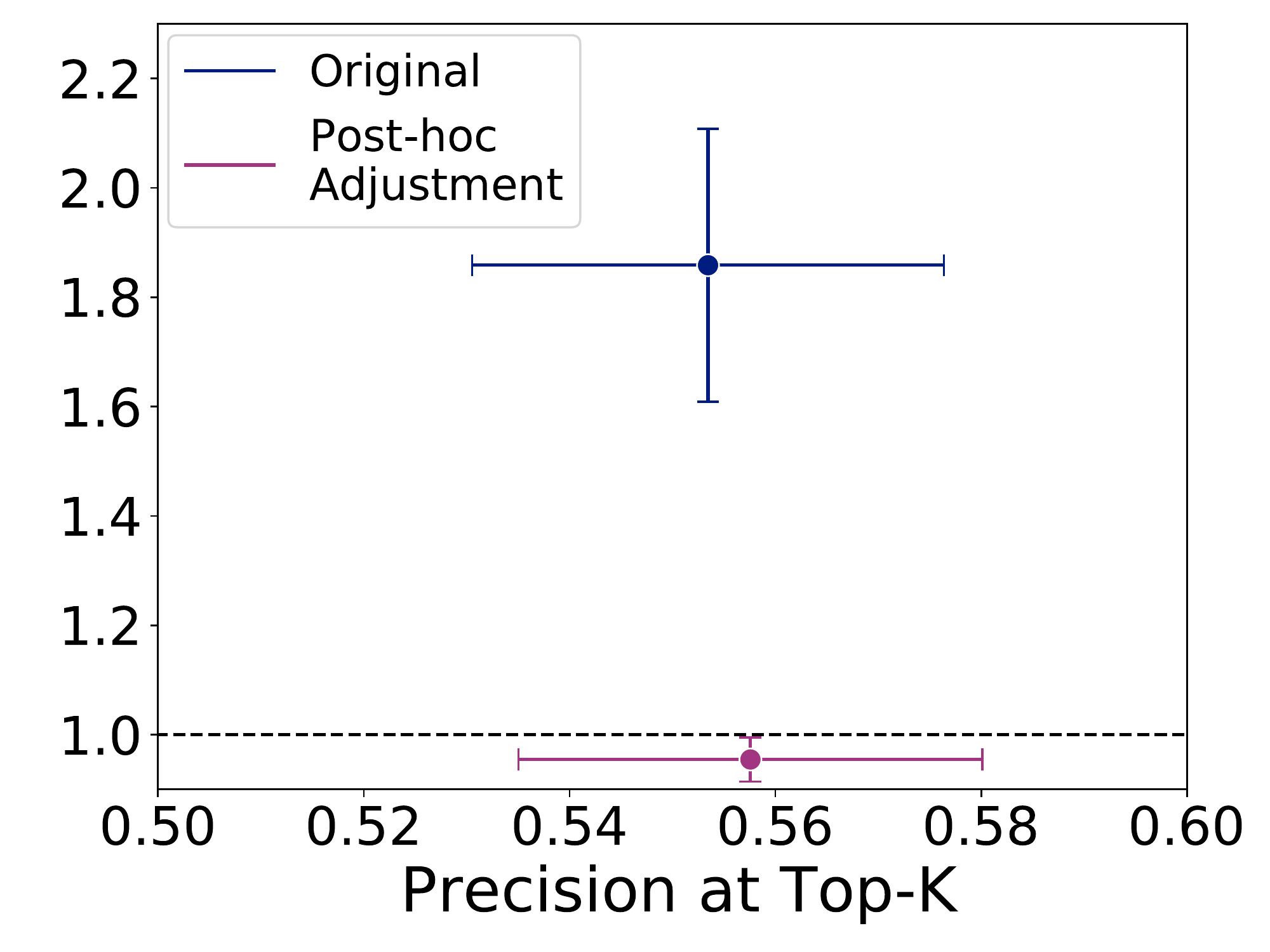}}%
    \caption{Education Crowdfunding}
   \end{subfigure}
   \caption{Effect of post-hoc adjustments on model accuracy (precision@k) and fairness (recall disparities). Separate score thresholds were chosen for the protected and non-protected subgroups to equalize recall across the groups. Error bars show 95\% confidence intervals over validation sets.}
   \label{fig:post_hoc}
\end{figure*}

Figure \ref{fig:post_hoc} shows the results of post-hoc adjustments to equalize TPR across groups by choosing separate, group-specific thresholds. Across all four policy settings, this approach consistently improved the fairness of the models, entirely removing the disparities in most cases. Notably, this improved fairness was achieved with negligible cost in terms of model accuracy in all four settings. While this lack of fairness-accuracy trade-off is somewhat surprising on its face, the ``top k'' setting likely plays a role here as well. With limited resources relative to needs, there are many ways to choose $k$ individuals for intervention with equally high precision, making it possible to swap some high-risk individuals from one group with those from another in order to improve fairness without appreciably reducing accuracy. To the extent that any small trade-offs may exist when making these adjustments, they seem to be dominated by variation over time in the generalization performance of the models, yielding consistently fair adjusted models without sacrificing accuracy (for a more detailed discussion of the lack of trade-offs with this approach, see our recent work in~\cite{rodolfa2020machine}).

\subsection{Effect of Composite Models}
\begin{figure*}[!hbtp]
  \centering
   \begin{subfigure}[b]{0.24\textwidth}
    \centering
    \raisebox{1mm}{\includegraphics[width=\textwidth]{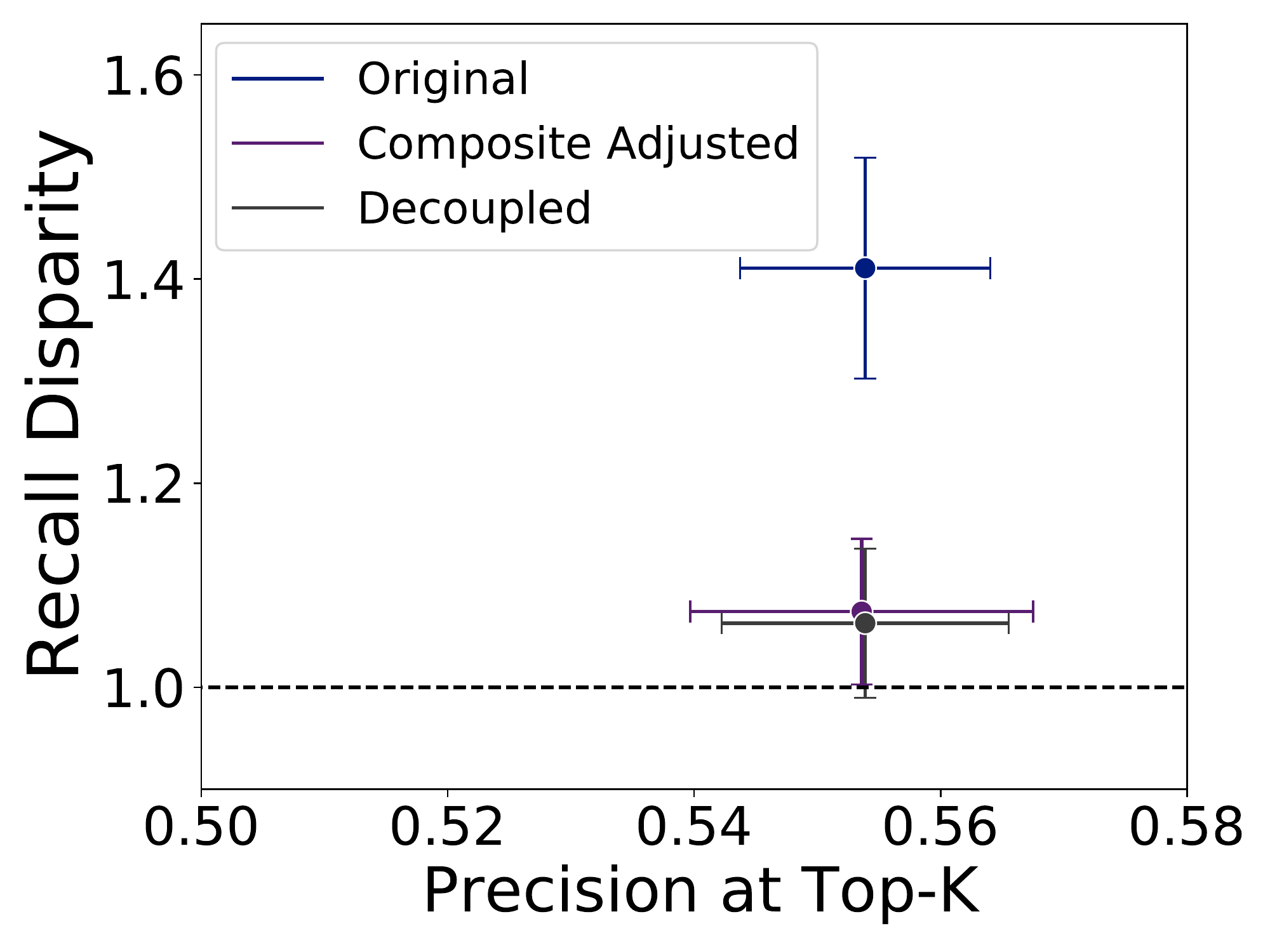}}%
    \caption{Inmate Mental Health}
   \end{subfigure}
   \begin{subfigure}[b]{0.24\textwidth}
    \centering
    \raisebox{1mm}{\includegraphics[width=\textwidth]{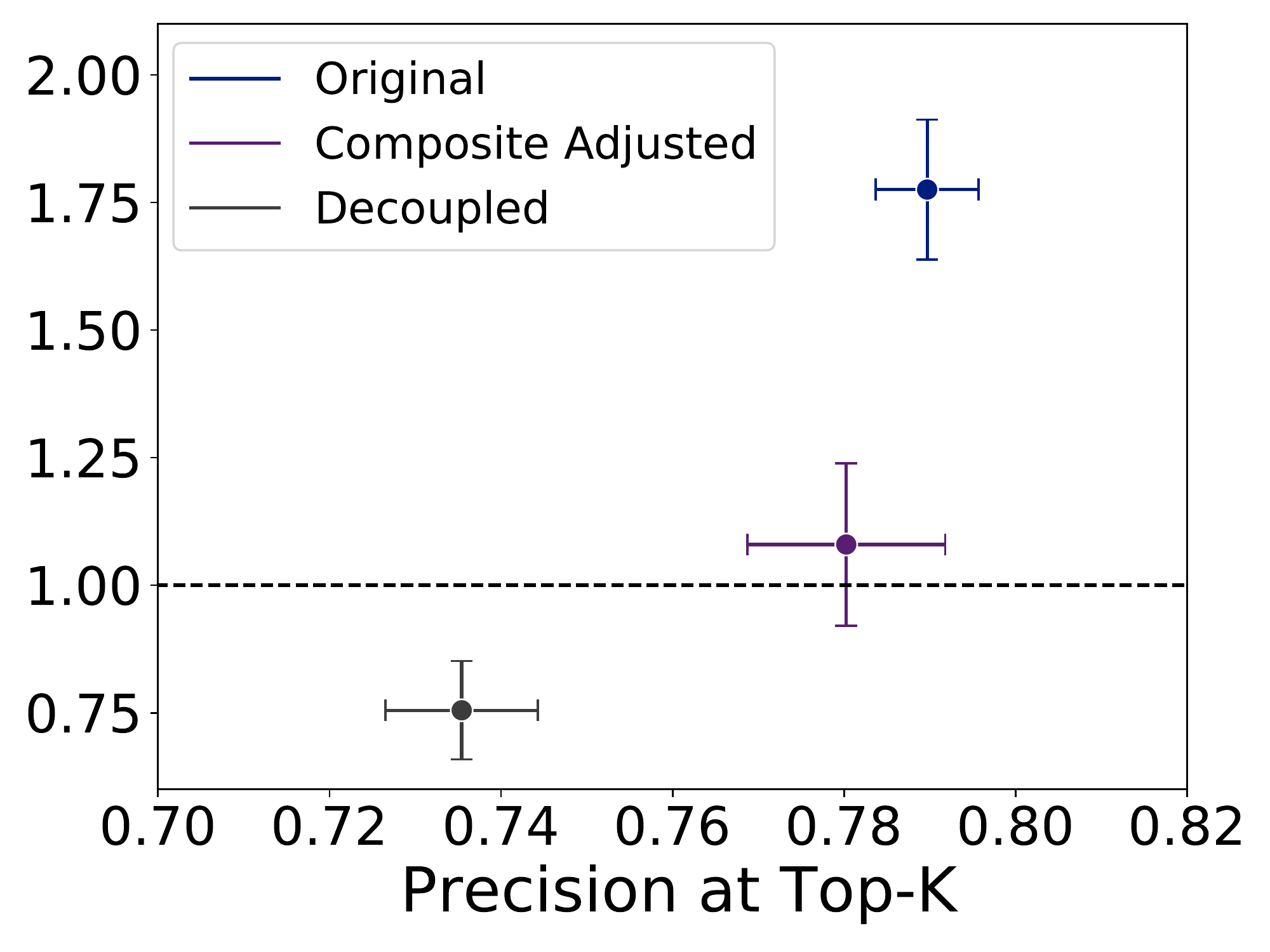}}%
    \caption{Housing Safety}
   \end{subfigure}
   \begin{subfigure}[b]{0.24\textwidth}
    \centering
    \raisebox{1mm}{\includegraphics[width=\textwidth]{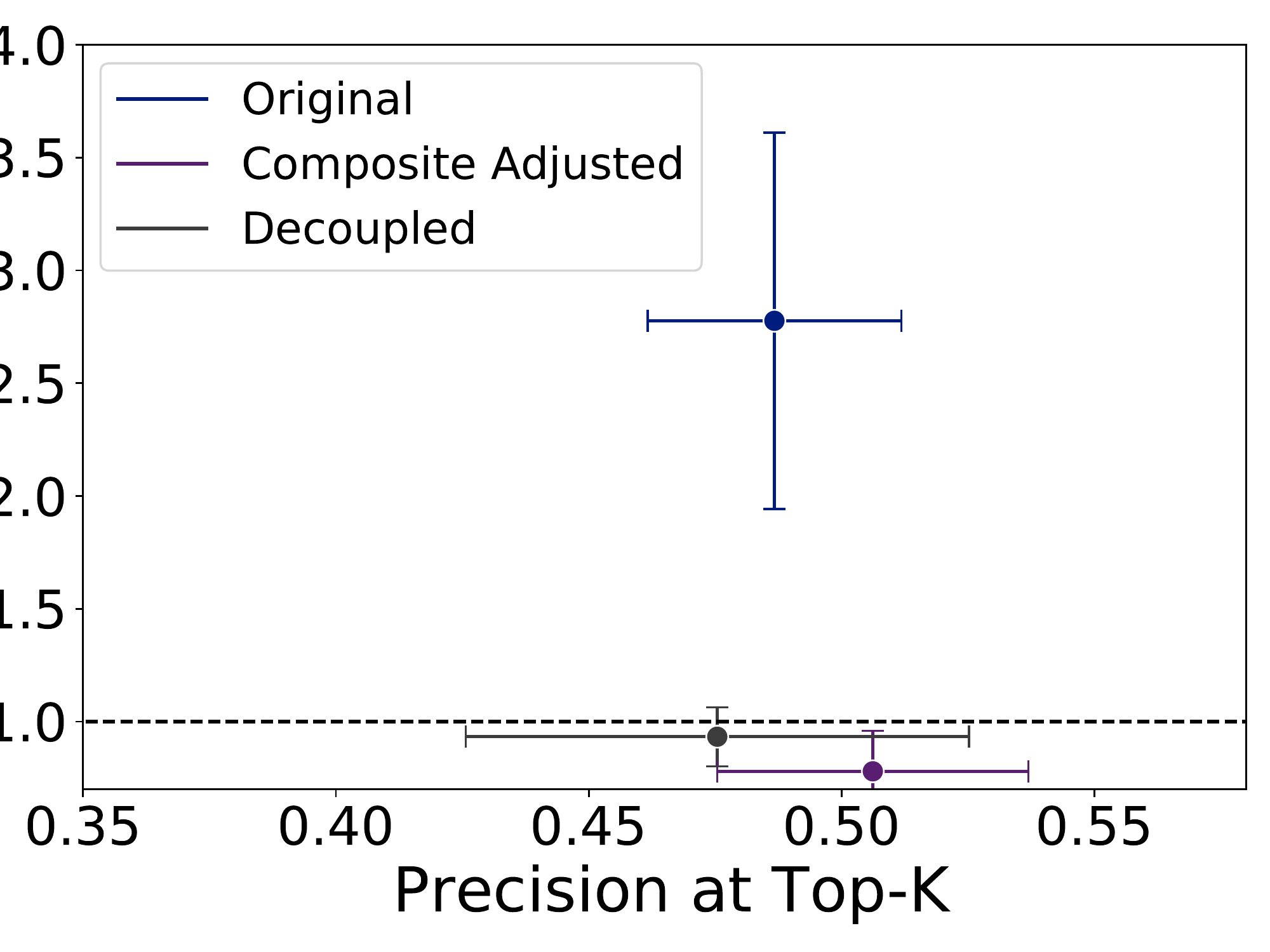}}%
    \caption{Student Outcomes}
   \end{subfigure}
   \begin{subfigure}[b]{0.24\textwidth}
    \centering
    \raisebox{1mm}{\includegraphics[width=\textwidth]{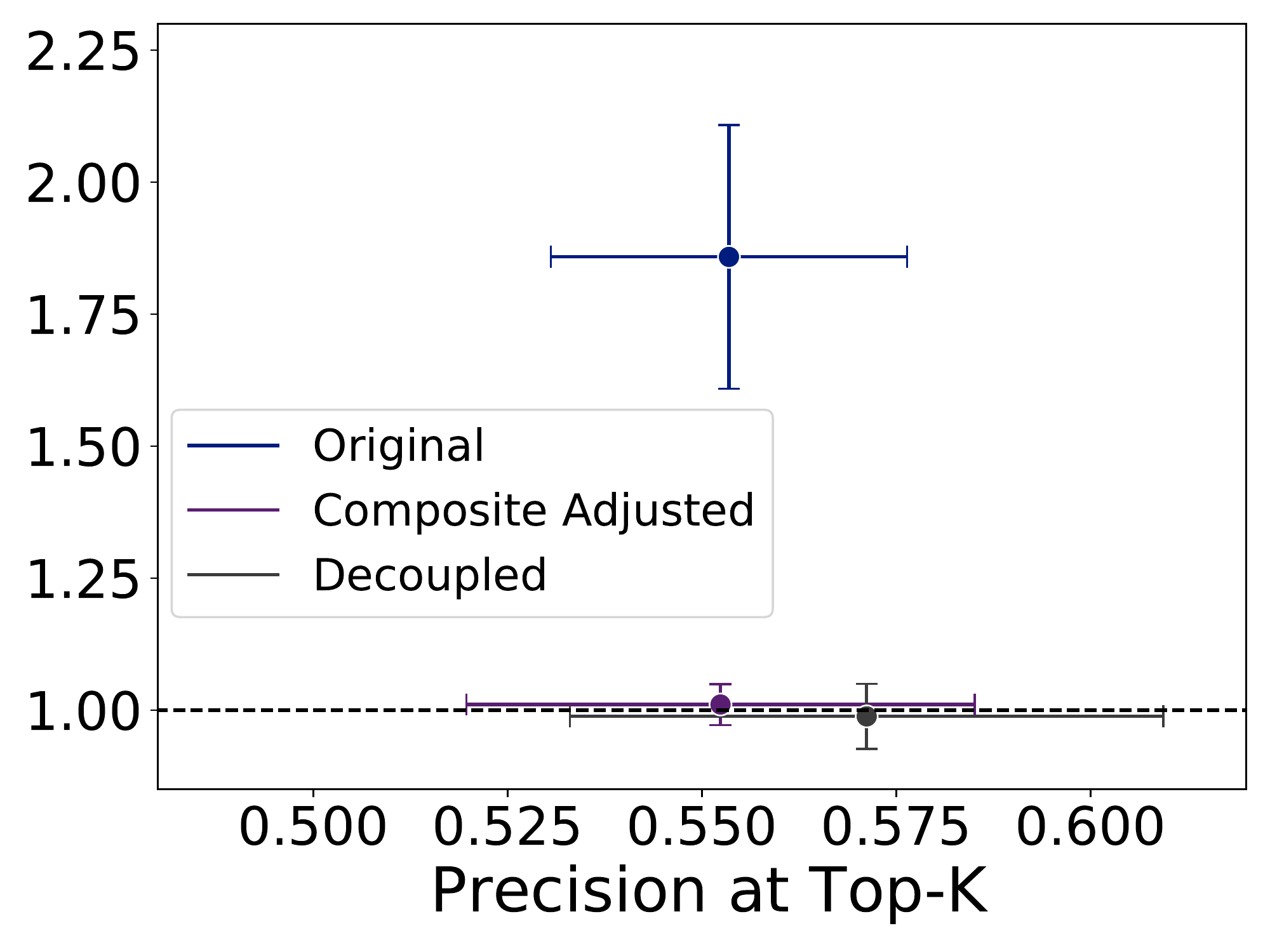}}%
    \caption{Education Crowdfunding}
   \end{subfigure}
   \caption{Effect of composite model approaches on model accuracy (precision@k) and fairness (recall disparities). Composite models were developed by choosing separate models for each subgroup either with or without decoupled training. Error bars show 95\% confidence intervals over validation sets.}
   \label{fig:composite}
\end{figure*}

The final approach explored here follows Dwork's proposal \cite{dwork2018decoupled} to build composite models, either through separate model selection or fully decoupled training for each subgroup. Figure \ref{fig:composite} presents the results of these two strategies in each of the policy settings. In general, we find these approaches to perform quite well, consistently reducing the disparities across all four settings. As noted above, because the uncalibrated scores of these group-specific models cannot be assumed to be comparable, we combined the models across groups by making use of the same process of choosing TPR-equalizing thresholds as we used to make post-hoc adjustments to single models. As such, the fairness improvements seen here might either be a result of the composite strategy itself or the method for choosing thresholds, which, as seen above was itself very successful in reducing disparities here. However, if selecting (or training) separate models was appreciably improving model performance for the subgroups, we might hope to see increases in the overall accuracy of the composite models relative to the post-hoc adjusted ones in Figure \ref{fig:post_hoc}, but the results here do not lend evidence to support this hypothesis. While there may be a slight improvement in precision@k for the composite model in the Student Outcomes setting, the difference in that setting is far from statistically significant and accuracy of the composite models in other settings is nearly identical to or somewhat lower than that of the post-hoc adjusted models.

To disentangle the effects of the composite modeling strategy itself from the TPR-equalizing group-specific thresholds used here, other strategies for choosing and combining the models across subgroups could be explored, although these are complicated somewhat by the requirement of the ``top k'' setting here that a total of $k$ entities is selected across groups. The score threshold yielding the desired number of entities will vary considerably across pairs of models, so the appropriate cut-off at which to evaluate model performance for one subgroup depends on what models it will be combined with for other subgroups. As a preliminary experiment, we explored a simplified strategy in which we selected models for each subgroup based on their performance among the same number of highest-risk individuals that would have been selected from a single (non-composite) model. These group-specific models were combined and then the top $k$ individuals with the highest scores in the composite model were chosen with a single score threshold.\footnote{Note that this approach will likely yield a different number of individuals in each group than was in process of selecting the group-specific models, so the assumption being made here is that these differences will be small enough that the model performance among each subgroup will not depart appreciably from what was used during selection.} In these initial experiments, composite models with a single score threshold failed to improve on either the accuracy or fairness of the original models, lending support to the conclusion that the improvements observed in Figure \ref{fig:composite} are likely driven by the TPR-equalizing thresholds used to combine the models across subgroups.

In most of the problem settings considered here, the composite models with and without fully-decoupled training performed similarly, but the Housing Safety context provides a notable exception (Figure \ref{fig:composite}(c)). Although the composite approach performs well in this setting, the decoupled strategy shows a considerable loss in precision as well as over-shooting the necessary adjustment to achieve a fair result (ending up with bias in the opposite direction). Notably, the Housing Safety dataset is considerably smaller than the others used here, with an order of magnitude fewer entities than the next-largest setting. As observed in \cite{dwork2018decoupled}, one potential disadvantage to decoupled model training is that the smaller number of training examples might degrade model performance, particularly if there are common patterns in the data that could be learned across groups. We would, of course, expect this issue to be exacerbated as the overall number of available examples decreases. Likewise, performing model selection on relatively small subgroups might be prone to over-fitting, choosing an overly-optimistic model specification whose performance reverts to a lower mean when measuring generalization performance on a future validation set. Such over-optimistic performance estimates for one subgroup could also affect the recall-balancing thresholds chosen across groups, leading to relatively too many individuals being chosen from one group and yielding disparities in the final composite model as well.
              
\section{Discussion}
The goal of the current work was to build on the extensive recent work in algorithmic fairness by comparing how the wide variety of proposed fairness-enhancing approaches and methods perform in the context of real-world problems in high-stakes policy problems. While our aim was not to comprehensively include every existing approach, we sought to sample a wide range of techniques applied at different phases of the machine learning pipeline by pre-processing of the input data, in-processing during model training, and post-processing of trained models. Similarly, we focus here on resource-constrained assistive policy contexts where the optimization problem reflects a ``top k'' setting and we argue TPR disparities are an appropriate fairness metric (reflecting a concept of ``equality of opportunity'' as discussed in~\cite{hardt2016equality,rodolfa2020case}). While some of the results described here might not generalize beyond this problem setting, we note that it is very commonly encountered in high-stakes decisions across education~\cite{aguiar2015who,lakkaraju2015machine}, healthcare~\cite{potash2015predictive}, criminal justice~\cite{helsby2018early}, social services~\cite{Bauman2018ReducingInterventions}, as well as many other contexts \cite{kumar2018using,ye2019using,chouldechova2018case,athey2017beyond,glaeser2016crowdsourcing}, and has been the most common formulation encountered in the more than 100 projects we have been involved in applying machine learning to social good problems with government and non-profit partners.

In this setting, \textbf{pre-processing methods showed decidedly mixed and inconsistent results}, with both sampling and omitting the protected attribute improving fairness in some contexts but not others. This inconsistency is perhaps not entirely surprising given the wide range of potential contributors to disparities at any stage of the machine learning pipeline \cite{rodolfa2020book}, only some of which we might expect these pre-processing methods to address. Unfortunately, it seems unclear \textit{a priori} whether these strategies will be effective in a given context (or, with sampling, what approach will work), making them unreliable as a fairness-enhancing approach. 

Similarly, \textbf{we found little success with removing disparities through in-processing}, but, as noted above, existing methods to add fairness constraints in the process of model training seem particularly poorly suited to the ``top k'' setting. In principle, developing new in-processing methods better suited to the ``top k'' setting should be feasible, but poses particular technical challenges. As other work developing methods for this setting (without fairness constraints) has observed, the loss function in this setting is, in general, not only non-convex, but discontinuous (as disjoint regions of the parameter space yielding exactly $k$ predicted positives must be connected by regions yielding either more or fewer than $k$) \cite{liu2016transductive}. To our knowledge, no methods presently exist that seek to improve fairness through in-processing for ``top k'' models, but we believe this could be an interesting future research direction.

By contrast, we found \textbf{consistent success across the four problems with eliminating disparities using post-hoc methods.} Across all four policy settings considered here, these improvements in model fairness could be accomplished without a corresponding trade-off in accuracy. Although such trade-offs are often assumed to be an inherent aspect of reducing disparities in machine learning models~\cite{feldman2015certifying,zliobaite2015relation,calmon2017optimized} making this result somewhat surprising, the resource-constrained nature of these policy settings may contribute to the lack of trade-off as discussed above. Further, the consistency with which fair predictions could be obtained without cost to accuracy across the settings considered here may have important implications for policymakers and machine learning practitioners, reinforcing the moral imperative to ensure the fairness of models deployed in similar contexts (see our recent work in~\cite{rodolfa2020machine} for a more detailed discussion of these policy implications). 

Given the success of these post-hoc adjustments across models and settings, we also investigated whether applying these adjustments on top of the pre-processing and in-processing strategies explored here could remove any residual (or newly introduced) disparities from those methods. Consistently, post-hoc adjustment by choosing thresholds to equalize TPR yielded more fair results, even when applied in combination with other strategies that failed to improve fairness themselves. While this result suggests a robustness of this strategy, we did not observe any improvement in model performance by combining post-hoc adjustments with other strategies, so we do not see any advantage to doing so in practice. 

%\krcomment{Especially if we're short on space, might cut the discussion of post-hoc on top of other methods since I'm not sure how much it adds.}

Finally, we should note the importance of considering the broader context in which a machine learning model will be applied. While the work here has focused on improving the fairness of a model's predictions, doing so is only one step in the process of ensuring outcomes of the broader socio-technical system are themselves equitable. In most policy contexts, these models are deployed in a manner intended to inform the decision-making process of a human expert such as a doctor, case worker, or school administrator, rather than being fully autonomous. As such, fairness in a model's recommendations is not necessarily a guarantee that interventions will be allocated fairly, depending on how and when these humans in the loop follow or override them. Further, the interventions themselves may not be equally effective for everyone. For instance, additional after-school tutoring might be difficult to attend for students who have work or family obligations in the afternoons, or programs offered only in English might not effectively serve individuals for whom it is not their first language. Likewise, when the labels themselves are measured in inaccurate and disparate ways, such as using arrests as a proxy for crime commission \cite{Angwin2016MachineBias,Chouldechova2017FairInstruments,Mayson2019BiasOut,Kroll2016AccountableAlgorithms,Harcourt2015RiskAssessment}, measures of fairness that take these labels as ``ground truth'' will fail to capture these underlying disparities. Understanding the implications for fairness at each stage of the process --- from label definition through modeling to decisions and interventions --- is essential to understanding and mitigating biases in deployed machine learning systems that impact people's lives. The work here explores one key aspect of this process, but machine learning practitioners and the policymakers who deploy and act on the systems they build must be cognizant of these broader contextual aspects as well.

% why using fairlearn is tricky (We can move that earlier)
% \subsection{relationship with equity and bias in outcomes}
% While the focus of this paper has  on comparing different approaches that reduce bias in the predictions of a model, it is important to note that doing that is distinct from achieving equitable outcomes for people being affected by that model. The ML model is one component in the entire process and while it is useful and important to focus on fairness of ML models, it is more important to improve and ensure fairness of the overall socio-technical system the model is a part of.

% Another important consideration is that all of the methods we cover here are taking a ``supervised learning'' view of fairness, where a true, ground truth, label exists and we can compute disparities in predictions compared against the label. In many cases, that  ground truth label is not objective but a reflection of the historical context in which it happened. For example, ...police or education... 

\section{Summary and Future Work}
In the present study, we explored the performance of several proposed fairness-enhancing methods on reducing bias and enhancing fairness in general and improving equality of opportunity (as measured by TPR disparities) in particular across four real-world policy contexts. Among the methods considered, we found that post-hoc adjustments to model scores by choosing TPR-equalizing group-specific score thresholds was capable of removing disparities without loss of accuracy in all four settings. Most directly, our results have implications for practitioners building and deploying machine learning systems in similar resource-constrained policy contexts for whom this post-hoc approach should be both straightforward to implement and likely to improve the fairness of their models. For the machine learning research community, we believe this work highlights the importance of evaluating new methods on real-world problems, in particular demonstrating a gap with how well-suited current in-processing methods are to this ``top k'' setting.

Although we focus here on characteristics of machine learning problems commonly encountered in high-stakes policy contexts, it will be important extend this work to other policy settings, particularly those for which other bias metrics beyond TPR disparity are of interest. In particular, we hope to understand whether the consistent improvements of the post-hoc adjustments employed here will generalize to other fairness metrics, especially those which are not guaranteed to be monotonically increasing or decreasing with the model score. Additionally, in all the settings considered here, the sensitive attribute was known exactly, but this is not always the case. Unfortunately, many of the approaches considered in this study (such as sampling, composite models, and post-hoc adjustment) cannot be directly applied where there is uncertainty around group membership, and more work will be required to both extend these methods to those contexts as well investigate the performance of methods that are inherently better-suited to them (such as those described in~\cite{celis2019classification,menon2018cost}). Finally, although we sought to sample a range of fairness-enhancing methods across pre-, in-, and post-processing approaches, many more methods have been proposed than we could incorporate in the present work and continuing to extend upon these findings with additional methods will be an interesting avenue for future work. 

%\rgcomment{perhaps add counterfactuals method here: We also omit counterfactual based methods since they generally assume an underlying causal model working similarly for all different protected groups and not accounting for the differences that might exist while generating counterfactual data points.}

% - other types of bias metrics beyond recall disparity
% - contexts where protected/non-protected group membership is not known exactly
% - extending to more methods
% - 

\section{Acknowledgements}
This project was partially funded by the C3.AI Digital Transformations Institute. We would also like to thank the Data Science for Social Good Fellowship fellows, project partners, and funders as well as our colleagues at the Center for Data Science and Public Policy at University of Chicago for the initial work on projects that were extended and used in this study.

\balance

\small{
\bibliographystyle{abbrv}
\bibliography{related_work_refs, kit_mendeley}
}

\clearpage

%\section{Appendix}
%\input{appendix}

\end{document}